%% file: main.tex
\documentclass[final,5p,times,twocolumn]{elsarticle}
\usepackage{amssymb}
\usepackage{amsmath,amsfonts}
\usepackage{algorithmic}
\usepackage{graphicx}
\usepackage{textcomp}
\usepackage{soul}
\usepackage{xcolor}
\soulregister{\ref}{7}
\soulregister{\cite}{7}
\usepackage{algorithm}
\usepackage{algorithmic}
\usepackage{subfigure}
\usepackage{epstopdf}
\usepackage{enumerate}
\usepackage{tabularx}
\usepackage{amsthm}
\usepackage{booktabs}
\usepackage{balance}
\usepackage{xspace}
\usepackage{multirow}
\usepackage{ulem}
\usepackage{array}
\usepackage{tabularx}
\usepackage{threeparttable}
\usepackage{caption}
\usepackage{makecell}
\usepackage{diagbox}

\newcommand{\Methods}[2]{\begin{tabular}[c]{@{}c@{}}{#1}\\ {#2}\end{tabular}}
\newcolumntype{L}[1]{>{\raggedright\arraybackslash}p{#1}}
\newcolumntype{C}[1]{>{\centering\arraybackslash}p{#1}}
\newcolumntype{R}[1]{>{\raggedleft\arraybackslash}p{#1}}

\usepackage{url}

\journal{Knowledge-Based Systems}

\begin{document}
\begin{frontmatter}

\title{SAM-LAD: Segment Anything Model Meets Zero-Shot Logic Anomaly Detection}

%% REMOVE EAD to drop email addresses of authors
\author[label1]{Yun Peng}\ead{pengyun@tongji.edu.cn}
\author[label1]{Xiao Lin}
\author[label1]{Nachuan Ma}
\author[label1]{Jiayuan Du}
\author[label1]{Chuangwei Liu}
\author[label1]{Chengju Liu\corref{cor}}\ead{liuchengju@tongji.edu.cn}
\author[label1]{Qijun Chen\corref{cor}}\ead{qjchen@tongji.edu.cn}

\affiliation[label1]{organization= {Shanghai Institute of Intelligent Science and Technology, College of Electronics and Information Engineering, Tongji University},
            city={Shanghai},
            country={China}}

\cortext[cor]{Chengju Liu and Qijun Chen are the corresponding author.}

\begin{abstract}
Visual anomaly detection is vital in real-world applications, such as industrial defect detection and medical diagnosis. However, most existing methods focus on local structural anomalies and fail to detect higher-level functional anomalies under logical conditions. Although recent studies have explored logical anomaly detection, they can only address simple anomalies like missing or addition and show poor generalizability due to being heavily data-driven. To fill this gap, we propose SAM-LAD, a zero-shot, plug-and-play framework for anomaly detection in any scene. First, we obtain a query image's feature map using a pre-trained backbone. Simultaneously, we retrieve the reference images and their corresponding feature maps via the nearest neighbor search. Then, we introduce the Segment Anything Model (SAM) to obtain object masks of the query and reference images. Each object mask is multiplied by the entire image's feature map to obtain object feature maps. Next, an Object Matching Model (OMM) is proposed to match objects in the query and reference images. To facilitate object matching, we propose a Dynamic Channel Graph Attention (DCGA) module, treating each object as a keypoint and converting its feature maps into feature vectors. Finally, based on the object matching relations, an Anomaly Measurement Model (AMM) is proposed to detect objects with logical anomalies. Structural anomalies in the objects can also be detected. We validate our proposed SAM-LAD using various benchmarks, including industrial datasets (MVTec Loco AD, MVTec AD), and the logical dataset (DigitAnatomy). Extensive experimental results demonstrate that SAM-LAD outperforms existing SoTA methods, particularly in detecting logical anomalies.
\end{abstract}


\end{frontmatter}

%% \linenumbers

%% main text
\input{content.tex}

\bibliographystyle{elsarticle-num} 
\bibliography{reference.bib}

\end{document}

%% file: content.tex
\section{Introduction}\label{sec:intro}

\begin{figure}[htbp]
	\begin{center}
		\centering
		\includegraphics[width=\linewidth]{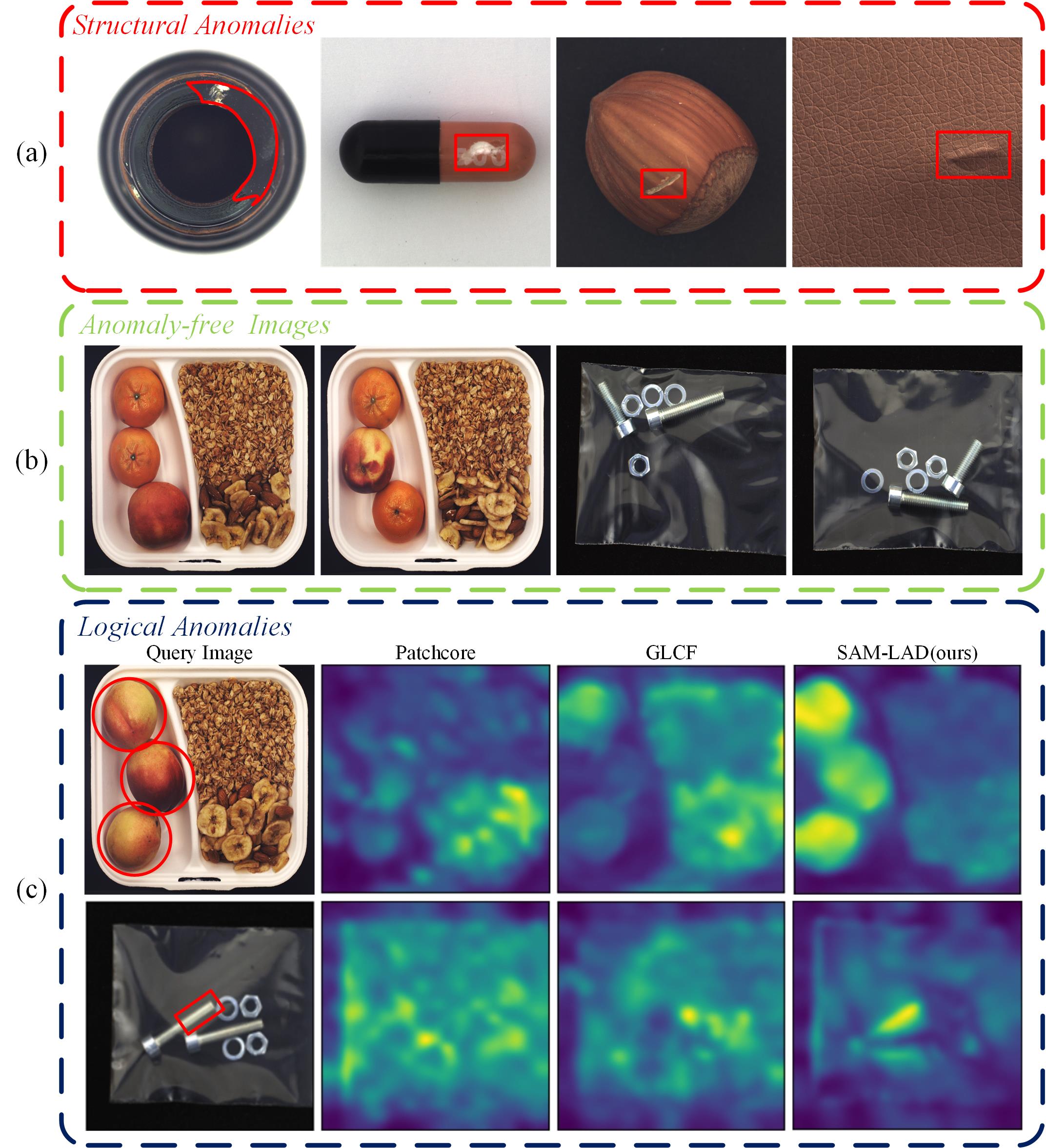}
		\centering
		\caption{(a) Example of the structural anomalies. (b) The anomaly-free images of the category breakfast box and screw bag. (c) The anomaly score map of the Patchcore, GLCF, and SAM-LAD for logical anomaly detection.}
		\label{1_1}
	\end{center}
\end{figure}

In recent years, image anomaly detection techniques have been widely applied in industrial quality detection\cite{bergmann2019mvtec}\cite{lyu2024reb}\cite{bergmann2020uninformed}\cite{tan2024unsupervised}, anomaly segmentation\cite{wei2024few}, and medical diagnosis scenarios\cite{li2024adapting}\cite{li2018thoracic}, aiming to detect abnormal data that are different from normal data within a sample image\cite{golan2018deep}. Since abnormal prior information is scarce, this task is commonly conducted within the framework of an unsupervised learning paradigm. Consequently, there has been increasing interest among scholars in researching unsupervised anomaly detection, and near-perfect results have been achieved, as evidenced by methods such as Pull\&Push\cite{zhou2022pull}, PMB-AE\cite{xing2023visual}, and Patchcore\cite{roth2022towards}. However, due to the limits of those scenarios, most anomaly detection methods currently focus on structural anomaly detection and can only deal with even one object in a single scene at a time, as illustrated in Fig.\ref{1_1}(a).

In many usual scenarios such as autonomous driving and surveillance systems, understanding the semantic context of the entire scene and detecting anomalies is essential. Therefore, Bergmann et al.\cite{bergmann2022beyond} have proposed logical anomalies, which represent more complex abnormalities in the logical relationships between objects within the entire scene. Logical anomalies violate underlying constraints, thereby contravening specific relationships between objects, e.g., a permissible object being present in an invalid location or a required object not being present at all. Specifically, within a scene, anomalies that appear normal at the local level but violate geometric constraints or logical principles when considering global semantics are referred to as global logical anomalies. For example, in Fig.\ref{1_1}(b), anomaly-free images of the category breakfast box always contain exactly two tangerines and one nectarine that are always located on the left-hand side of the box. Furthermore, the ratio and relative position of the cereals and the mix of banana chips and almonds on the right-hand side are fixed. An anomaly-free screw bag contains exactly two washers, two nuts, one long screw, and one short screw. Nevertheless, their logical anomalies could involve missing, extra, wrong location, wrong combination, etc.

For the more challenging global logical anomalies, most existing methods tailored for structural anomalies demonstrate catastrophic results, for instance, as illustrated in Fig.\ref{1_1}(c) with Patchcore. This is because structural or textural anomalies belong to lower-level anomaly types, which do not necessitate understanding the overall semantics of objects in the scene and only require local knowledge for anomaly detection. However, for higher-level global logical anomalies, relying solely on local perception to ascertain the normalcy of overall semantics is insufficient. Consequently, the performance of existing methods is significantly constrained.

Our previous work\cite{peng2023semi}\cite{liu2022semi} proposed a reconstruction-based method for detecting multiple object anomalies to address logical anomalies. In \cite{peng2023semi}\cite{liu2022semi}, objects had fixed spatial relationships. We used semi-supervised learning on positive samples to regress object positions in the test samples and reconstruct their normal features. However, exploring new methods is essential to address logical anomalies in more complex scenarios like \cite{bergmann2022beyond}, where objects lack fixed spatial relationships.

Currently, Yao et al. proposed GLCF\cite{yao2023learning}, Guo et al. proposed THFR\cite{guo2023template}, and Zhang et al. proposed DSKD\cite{zhang2024contextual}. These state-of-the-art methods, based on reconstruction and knowledge distillation, achieve decent performance in detecting logical anomalies. For example, GLCF attempted to focus on anomalies in nectarines on the left-top side, as shown in Fig.\ref{1_1}(c). Unfortunately, GLCF exhibited disappointing results when encountering the screw bag. This is because, in the screw bag scenario, anomaly-free samples involve randomly placed objects, which requires a high level of contextual understanding and the exclusion of interference from diverse positive sample features during inference. Additionally, these methods are heavily data-driven, making them poorly generalizable and costly to retrain for new scenarios.

To address the above challenges: (1) the presence of multiple key objects in complex scenes, (2) variability in positive sample features, and (3) poor generalization due to dependence on data-driven methods. We propose a novel framework called SAM-LAD, which can be plug-and-play in any scenario without training and even outperforms existing data-driven logical anomaly detection methods. Specifically, we ingeniously leverage the robust object segmentation capability of the Segment Anything Model (SAM)\cite{kirillov2023segment} to obtain the positional information of all key objects in the scene by setting segmentation thresholds. Then, we utilize a pre-trained DINOv2\cite{oquab2023dinov2} backbone network as a feature extractor to extract features and employ nearest neighbor search to find the $k$ most similar normal samples to the query image. Following this, we implement the FeatUp\cite{fu2024featup} operation to upsample the feature maps and recover their lost spatial information. Combining the upsampled feature maps with the positional information obtained from SAM yields separate feature maps for each object. Subsequently, we propose a Dynamic Channel Graph Attention (DCGA) mechanism to effectively compress each object's feature map into a single feature vector. Consequently, leveraging the proposed Object Matching Model (OMM), we match the feature vectors of the reference images with those of each feature vector in the query image. Finally, based on the matching results, we propose an Anomaly Measurement Model (AMM), which estimates the feature distribution of individual objects in the query image and the corresponding $k$ matched objects from the $k$ normal samples, thereby calculating the final anomaly score map. Noteworthy, thanks to the introduction of AMM, our framework not only excels in detecting global logical anomalies but also fulfills the requirements for detecting structural anomalies. Our framework is motivated by the following: when humans discern anomalies among multiple objects within intricate scenes, the most straightforward and most efficacious approach involves individually juxtaposing several normal samples with each object in the test sample. Through this meticulous comparison, inference regarding anomalous regions can be inferred without necessitating the establishment of a costly global semantic contextual comprehension within the model. The essence lies in identifying and aligning pivotal objects, thus accomplishing the entirety of the task. We evaluate the proposed framework on multiple commonly used benchmarks, and the experimental results demonstrate that our SAM-LAD achieves state-of-the-art performance. The main contributions of this paper can be summarized as follows:

\begin{enumerate}
    \item We propose the SAM-LAD to address the challenge of detecting logical anomalies. This framework introduces an object-level matching algorithm to determine the correspondence between objects and normal images. Based on the correspondence, a statistical estimator is designed to compute the feature estimation of the object. By analyzing the feature estimation differences between paired objects, we can detect logical and structural anomalies simultaneously, improving the performance of visual anomaly detection models.
    \item We integrate the visual large model SAM into logical anomaly detection and leveraged its powerful generalization capabilities to achieve zero-shot logical anomaly detection without additional training.
    \item We conducted experiments on multiple benchmarks, showcasing the state-of-the-art (SoTA) performance of our method.
\end{enumerate}

\section{Related Work}\label{sec:rw}

\subsection{Semantic Segmentation}
Semantic segmentation has seen significant advancements, with novel methods targeting improved precision and versatility. The Segment Anything Model (SAM)\cite{kirillov2023segment} revolutionizes object segmentation by leveraging large training datasets and innovative architectures. Existing studies have explored techniques like transformers with self-attention\cite{vaswani2017attention} and methods that integrate diverse image contexts to enhance segmentation. Before SAM, models like Mask R-CNN\cite{he2017mask} and DeepLab\cite{chen2017deeplab} extended segmentation capabilities using region proposals, feature pyramid networks, and atrous convolutions. SAM continues this innovation with a prompt-driven architecture that enhances traditional segmentation tasks and explores new segmentation capabilities. Interactive segmentation methods have addressed object boundary ambiguity by enabling user inputs to guide predictions. However, SAM differs by providing flexible segmentation prompts (e.g., points, boxes, masks) and generalizing segmentation to diverse scenarios.

In this work, we leverage SAM's robust segmentation capabilities and generalization performance. By setting segmentation thresholds, we effectively isolate objects from the scene. Thus, our framework achieves groundbreaking scene analysis based on zero-shot learning.

\subsection{Keypoint Matching}
Keypoint matching is essential for object recognition, 3D reconstruction, and image stitching in computer vision. The long-standing interest in this problem has led to various approaches. Early methods like SIFT\cite{lowe2004distinctive} and ORB\cite{rublee2011orb} laid the groundwork with handcrafted keypoint descriptors. These algorithms created robust descriptors invariant to scale, rotation, and partial occlusion, making them practical for many tasks despite sensitivity to image illumination changes. Recently, deep learning techniques revolutionized keypoint matching by learning discriminative features from large datasets. CNNs, like the SuperPoint model\cite{detone2018superpoint}, have been widely adopted for keypoint detection and matching, learning detection, and descriptor generation end-to-end. Deep learning approaches offer improved robustness and generalization over handcrafted methods. Recent advances use attention mechanisms to improve keypoint matching in challenging environments. Networks like SuperGlue\cite{sarlin2020superglue} use graph neural networks with self- and cross-attention to establish robust keypoint correspondences. Transformers in keypoint matching improve context understanding, handling complex matching scenarios more proficiently.

Inspired by SuperPoint and Superglue, we consider each object in the scene as a keypoint, and compress the keypoint features into a single feature vector using the proposed dynamic channel graph attention (DCGA) mechanism. Then, we establish object-to-object matching relationships using the proposed OMM.

\subsection{Unsupervised Anomaly Detection}
Current unsupervised anomaly detection approaches are typically classified into two categories: methods targeting structural anomalies and those targeting logical anomalies. Most current anomaly detection methods target structural anomalies, using robust feature extraction networks to obtain high-level semantic features of a test image. The distance between features of test images and anomaly-free images is calculated, identifying areas with large distances as anomalies.
Existing methods typically use deep CNN models like ResNet\cite{he2016deep} and EfficientNet\cite{tan2019efficientnet} for feature extraction. For example, Roth et al.\cite{roth2022towards} proposed Patchcore to detect anomalies on objects with a concise background. They selected ResNet-50\cite{he2016deep} pretrained on ImageNet\cite{krizhevsky2012imagenet} as the feature extractor. However, these methods struggle with logical anomalies since the local structures within the image appear normal. Therefore, more researchers are focusing on logical anomaly detection. For instance, Zhang et al. proposed DSKD\cite{zhang2024contextual}, and Batzner introduced EfficientAD\cite{batzner2024efficientad}. Both methods use the Student-Teacher network, detecting anomalies by comparing teacher and student differences. Using a local-global branching approach, Yao et al. presented GLCF\cite{yao2023learning}, while Bergmann et al. proposed GCAD\cite{bergmann2022beyond} to understand the logical semantic constraints of the entire image scene. Guo et al. presented THFR\cite{guo2023template} based on template-guided reconstruction. These methods have shown promise in detecting certain logical anomalies, such as missing components and unexpected excess. However, their performance drops significantly with more complex logical anomalies, such as misordering, mismatches, and haphazard object arrangements.

The proposed SAM-LAD builds a zero-shot method without strenuous efforts to comprehend the logical semantics of the entire scene. By introducing SAM and proposing OMM and AMM, a global logical anomaly detection system was developed, effectively addressing the limitations of the existing methods mentioned above.

\section{Proposed methodology}\label{sec:model}
\begin{figure*}[!t]
	\centering
	\includegraphics[width=\textwidth]{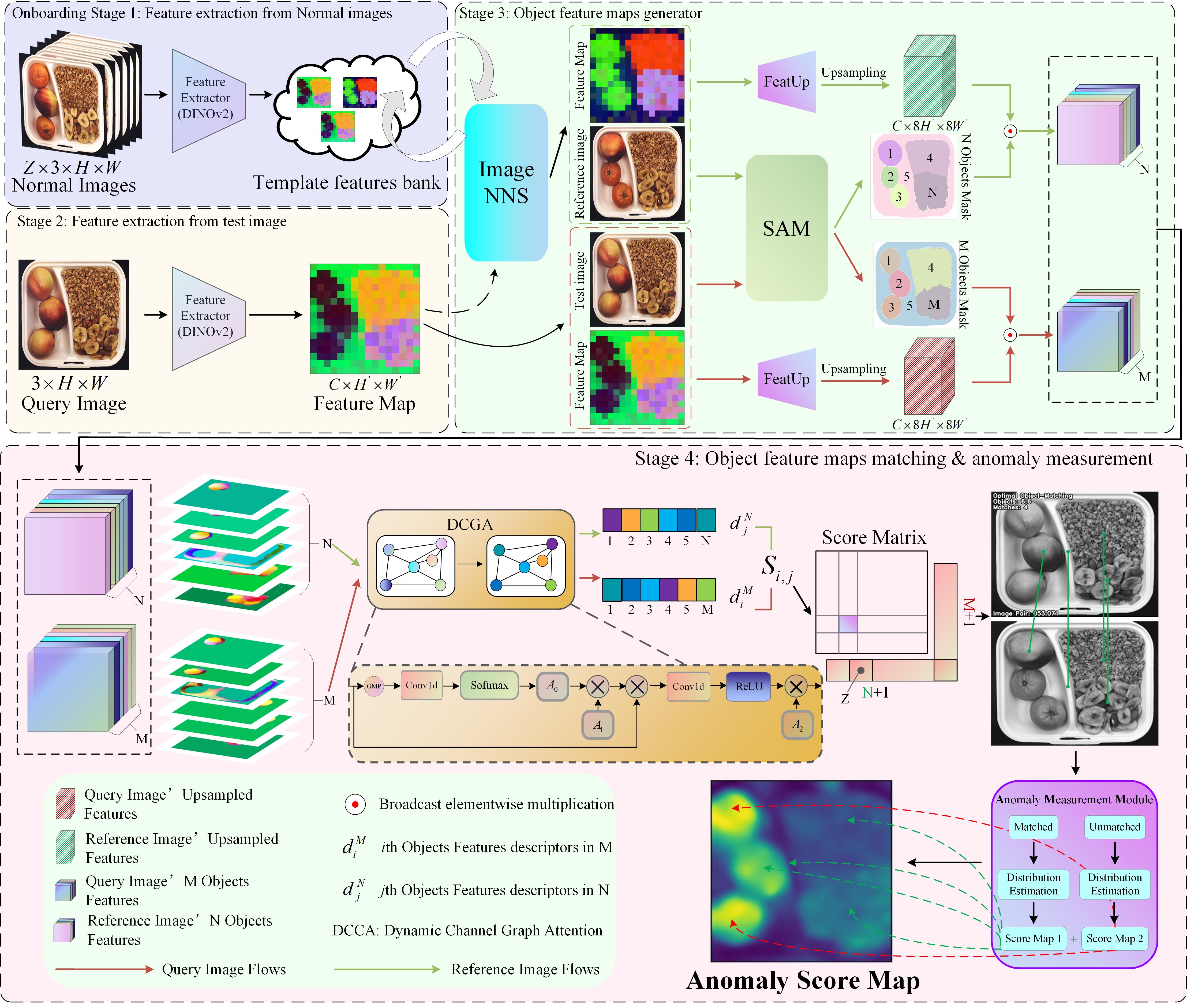}
	\caption{Pipeline of the proposed framework, which consists of four stages. The first stage is an offline operation, building an anomaly-free template features bank. The second stage is extracting a feature map from a query image. The next stage utilizes SAM to obtain object feature maps further. The last stage involves matching the objects in the query image with those in the reference images one by one. We calculate anomalies within each object and obtain the final anomaly score maps using the obtained matching relationships.}
	\label{framework}
\end{figure*}

\subsection{Architecture Overview}
The data flow of the proposed SAM-LAD is depicted in Fig.\ref{framework}, which consists of four stages. 1) On-boarding stage 1 is an offline operation. For all normal images $\mathbb{R}^{3 \times H \times W}$, a pre-trained backbone as a feature extractor is first deployed to extract the features from all normal images. Subsequently, the extracted features are compiled and stored within a template features bank. 2) Stage 2 involves extracting features from a query image $I_{q} \in \mathbb{R}^{3 \times H \times W}$ using the same feature extractor as stage 1. 3) Stage 3, for the feature maps $f_{q}$ extracted from a query image $I_{q}$, we retrieve its $k$-nearest normal feature maps $(f_{r}^{i})_{i\in{[1,k]}}$ in the template bank and their corresponding images $(I_{r}^{i})_{i\in{[1,k]}}$. We designate the $k$ normal images procured as reference images and in parallel with the query image, input $(I_{q}, I_{r}^{i})_{i\in{[1,k]}}$ into the SAM to obtain the individual object mask. Concurrently, the feature maps $(f_{q}, f_{r}^{i})_{i\in{[1,k]}}$ are subjected to a FeatUp operation, upsampling to 8$\times$. By combining the individual object masks with the upsampled feature maps, we obtain the object feature maps for both the query and reference images, respectively. 4) In Stage 4, the DCGA module is employed to encode and compress the object feature maps into object descriptor vectors. Subsequently, utilizing the proposed Object Matching Module, the two sets of object descriptor vectors are matched. Based on the matching results, the Anomaly Measurement Module computed the ultimate anomaly score map.

\subsection{Feature Extraction and Template Features Bank}

The first and second stage of the proposed SAM-LAD is the extraction of feature maps. The same feature maps are later used for FeatUp operation. There are multiple options for extracting features. Recently, Vision Transformer (ViT) has exhibited remarkable performance in anomaly detection-related tasks\cite{li2024promptad} \cite{zhang2024feature} due to its self-attention mechanism, enabling the model to attend to global information of the entire image when processing image patches. Therefore, we have adopted the pre-trained ViT-type DINOv2-S backbone\cite{oquab2023dinov2} as the feature extractor for the proposed SAM-LAD. For a given image $I_{x}^{3 \times H \times W}$, we denote the extracted feature maps $f_{x}^{C \times H^{'} \times W^{'}}$:
\begin{equation}
    f_{x} = F(I_{x}),
\end{equation}
where $F(\cdot)$ is DINOv2 feature extractor.

At initialization, we execute offline operations to construct a template features bank $\mathcal{B} = \left\{B_{1},  \cdots, B_{i}, \cdots, B_{Z}, \right\}$ using $F$ to extract $Z$ normal images in the all normal set $\mathbb{R}^{3 \times H \times W}$, where $B_{i}$ represents the template feature map of the $i$-th normal sample. At inference, only the feature maps of the query image are extracted, and we could reduce the template features bank using coreset subsampling method\cite{agarwal2005geometric}\cite{clarkson2010coresets} to reduce inference time and memory usage.

\subsection{Image-level Nearest Neighbor Search}
Given a query image $I_{q} \in \mathbb{R}^{3 \times H \times W}$, we take the extracted feature map $f_{q}$ as the query to retrieve its correlated template. ImageNNS obtains the template with index $t$ by randomly selecting from the template candidates that are $k$ most similar templates to increase the robustness during the inference process. The template selection process could be formulated as follows:
\begin{equation}
t=\operatorname{random}\left(\underset{\mathcal{S} \subset\{1, \cdots, Z\},|\mathcal{S}|=k}{\arg \min } \sum_{i \in \mathcal{S}} d\left(f_q, B_i\right)\right),
\end{equation}
where $d(\cdot)$ denotes the images-level distance between input query feature map $f_{q}$ and template feature key $B_{i}$ by flattening them to vectors to compute Euclidean metric. $S$ is a subset of $\left\{1, \cdots, Z\right\}$ denotes the indexes of $k$ template candidates, which are the top-k nearest of the input feature map $f_{q}$. Compared with the traditional point-by-point search strategy, imageNNS not only ensures that the reference is completely normal but also improves search efficiency.

Based on the obtained index $t$, we acquire the $k$ closest reference images $(I_{r}^{1}, I_{r}^{2}, \cdots, I_{r}^{k})^{k \times 3 \times H \times W}$ along with their corresponding feature maps $(f_{r}^{1}, f_{r}^{2}, \cdots, f_{r}^{k})^{k \times C \times H{'} \times W^{'}}$.

\subsection{Object Feature Map Generator}
\subsubsection{FeatUp Operation}
For a given query image $I_{q}$, we now possess $k$ pairs of feature maps $(f_{q},f_{r}^{i})_{i \in [1,k]}$. 
These deep feature maps already capture the semantics of the images. However, these feature maps lack spatial resolution, making them unsuitable for directly performing subsequent dense anomaly detection and segmentation tasks, as the model would aggressively pool information over large areas. Therefore, inspired by FeatUp\cite{fu2024featup}, we have employed its pre-trained feed-forward JBU upsampler\cite{fu2024featup} to restore the spatial information lost in the existing deep feature maps while still retaining the original semantics. Considering the balance of computational resources, inference speed, and ultimate detection accuracy, we opt to upsample the original feature maps by 8$\times$ (further analysis in Section \ref{section4.6}). Subsequently, we upsample each pair of feature maps to obtain the upsampled feature maps, which size is $(C \times 8H^{'} \times 8W^{'})$:
\begin{equation}
    ((f_{q})^{'},(f_{r}^{i})^{'}) = JBU (f_{q},f_{r}^{i}), i \in [1,k]
\end{equation}

\subsubsection{Object mask}
\begin{figure}[htbp]
	\centering
	\includegraphics[width=0.85\linewidth]{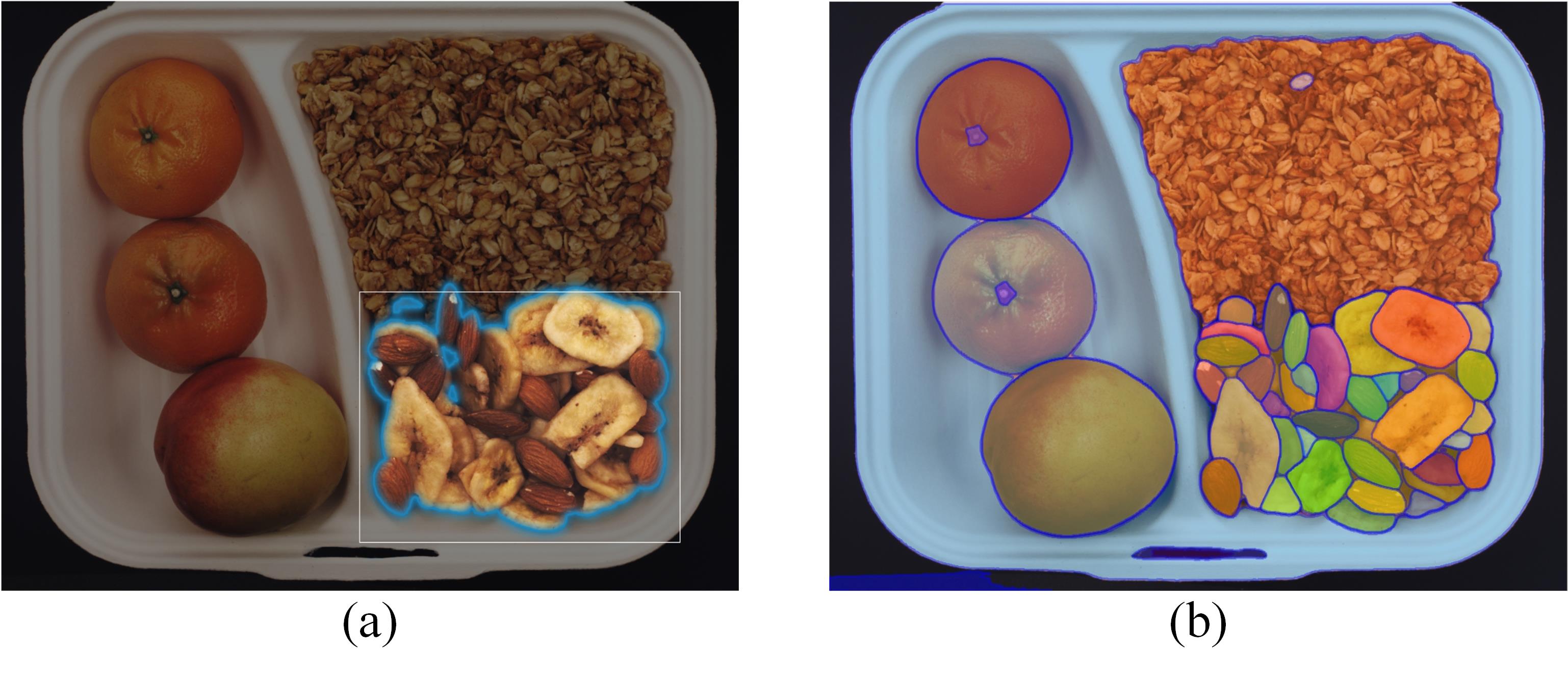}
	\caption{(a) A target object of breakfast box category, (b) Unexpected overly detailed objects mask.}
	\label{3_2}
\end{figure}

We use SAM to generate object masks for the input image pairs $(I_{q}, I_{r}^{i})_{i \in [1,k]}$. However, in actual applications, due to SAM's superior segmentation performance, it segments all potential objects in an image that are overly sensitive objects or overly generalized. For example, in Fig.\ref{3_2}(a), in the breakfast box category, a target object might be a mix of banana chips and almonds, but SAM segments individual banana chips and almonds(Fig.\ref{3_2}(b)), which is not what we intended. Thus, after obtaining the segmentation masks from SAM, it is necessary to filter them further to acquire the desired key object masks.

\begin{figure}[htbp]
	\centering
	\includegraphics[width=\linewidth]{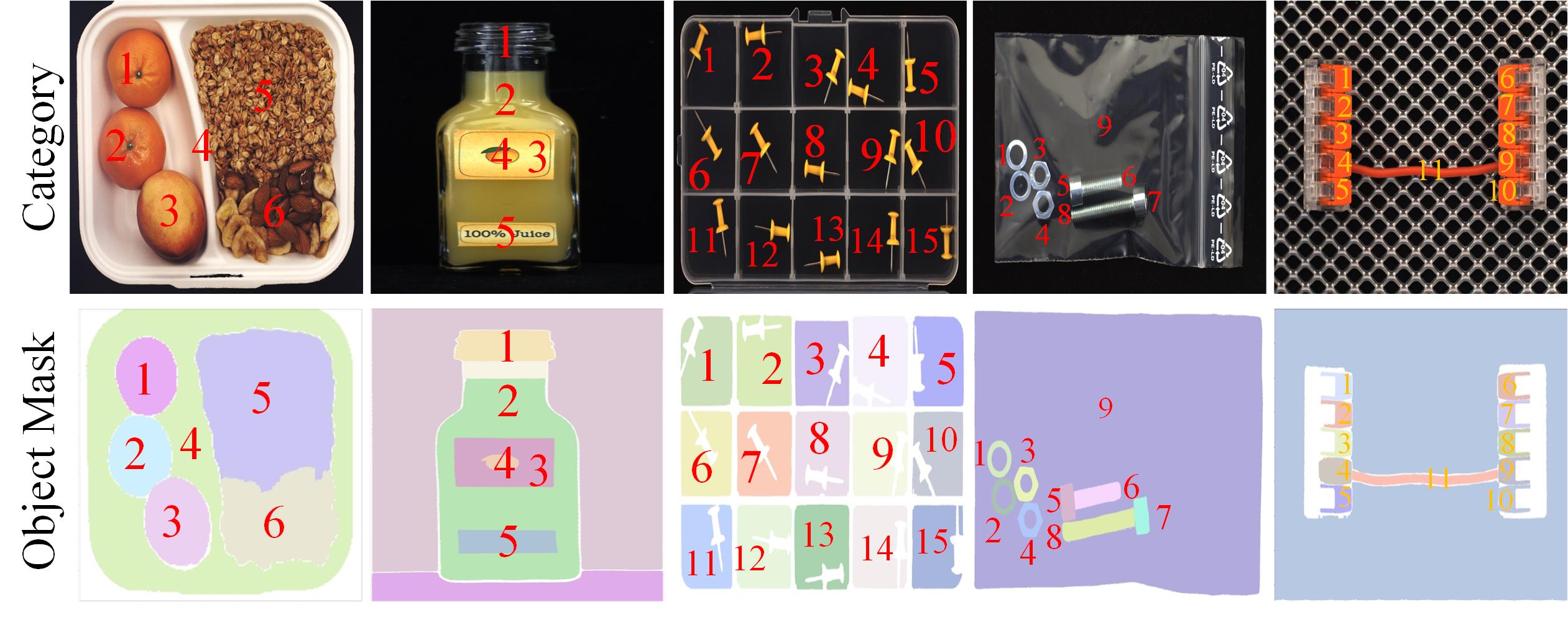}
	\caption{The anomaly-free image's object mask of each category dataset from SAM.}
	\label{3_3}
\end{figure}

Specifically, upon processing the image data through the SAM, we can obtain the area of each segmented object. By setting thresholds for the minimum and maximum area, we ignore objects that are not intended for detection and select the key objects of interest. For each scene category, SAM filters are set to determine the minimum and maximum area thresholds, resulting in object masks. The key object of interest is illustrated in Fig \ref{3_3}.

After segmentation by SAM and filtration, we obtain the $k$ pairs of segmented object masks for the image pairs $(I_{q}, I_{r}^{i})_{i \in [1,k]}$. For the segmented object mask from anomaly-free reference image $I_{r}$, the number of object classes within its object mask is invariably constant. For instance, the category of the breakfast box, an anomaly-free image always contains two tangerines, one peach, a portion of cereals, and a mix of banana chips and almonds, in addition to the main breakfast box itself, totaling six objects. Another category of screw bag contains exactly two washers, two nuts, one long screw, two screw heads, and one short screw, totaling eight objects. Hence, for the segmentation of object masks from the anomaly-free reference image $I_{r}$, we partition it into $N$ individual object masks. (For detailed $N$ values for each category, refer to the red numerical annotations of object masks in Fig.\ref{3_3}). For the segmented object masks from $I_{q}$, due to the presence of missing or additional objects, the number of object classes M in the segmented object masks varies. We partition it into M object masks. Ultimately, for the image pairs $(I_{q}, I_{r}^{i})_{i \in [1,k]}$, we obtain $k$ pairs of M and N object masks, respectively.

\subsubsection{Object Feature Map}
For the $M$ object masks from the query image $I_{q}$, we first use bilinear interpolation to scale them to a size of $(8H^{'}, 8W^{'})$. Then, we perform element-wise multiplication with the corresponding upsampled feature map $(f_{q})^{'}$, yielding $M$ object feature maps $(M, C, 8H^{'}, 8W^{'})$. Similarly, for the $N$ object masks from the reference image $I_{r}$, we perform the same operation to obtain $N$ object feature maps $(N, C, 8H^{'}, 8W^{'})$. Ultimately, we have obtained $k$ pairs of object feature maps $(f_{q}^{obj}, (f_{r}^{obj})^{i})_{i \in [1,k]}$. 

\subsection{Object Matching}
\subsubsection{Dynamic Channel Graph Attention}
\begin{figure}[htbp]
	\centering
	\includegraphics[scale=0.5]{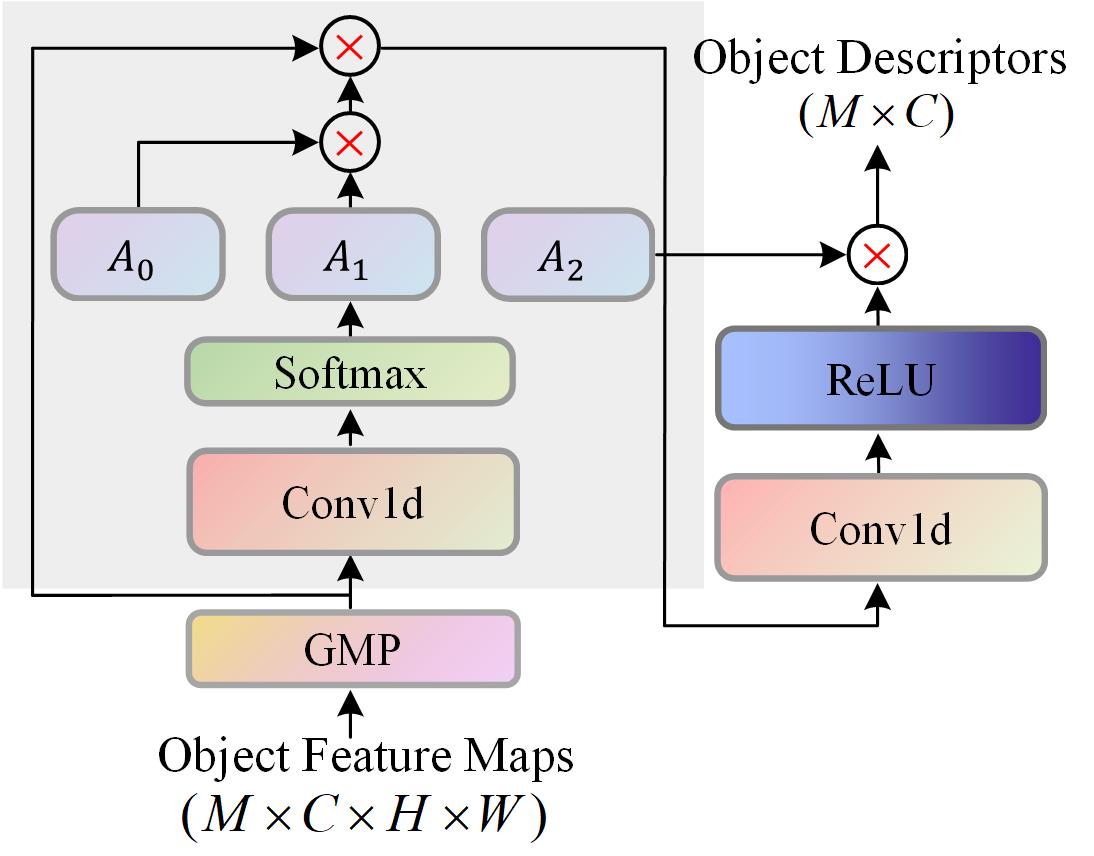}
	\caption{Architecture of the DCGA module.}
	\label{DCGA}
\end{figure}

To facilitate the matching of objects between the query and reference images, we consider each object to be a keypoint and extract their respective feature map into a feature descriptor vector. However, during the process of feature compression, it is inevitable that positional information will be lost. Additionally, the recalibration of channel weights by enhancing or suppressing semantic information renders the extraction of global information challenging. Therefore, the establishment of a Dynamic Channel Graph Attention (DCGA) module is proposed to enhance the responsiveness between objects and channels, which can explicitly capture the spatial dependencies of various objects to augment global representation. Specifically, the feature maps of all objects can be conceptualized as a graph structure. Within this concept, each individual object feature can be regarded as a vertex in the graph, and the relationships between these objects are seen as the edges. Moreover, within the feature map of each object, each individual channel can also be viewed as a vertex, with the interactions between these channels representing edges. A schematic of the DCGA mechanism is illustrated in Fig.\ref{DCGA}.

For a query image's objects feature map $(M,C,8H^{'},8W^{'})$, the dimension of each object feature map is first squeezed to $C \times 1 \times 1$ by a global max pooling (GMP) operation. Then, the object features $f_{\text {in }}$ in DCGA is a tensor of shape $M \times C$. Subsequently, a graph structure is employed to generate the weights for each vector.

This graph structure consists of two parts. The first part targets each object's feature vector $(1, C)$. Specifically, two independent $C \times C$ matrices, that is, $A_{0}$ and $A_{1}$, constitute the adjacency matrices, representing the dependency relationships among channel vertices. $A_{0}$ is a predefined identity matrix, representing only the vertex itself, and requires normalization. $A_{1}$ is a self-attention-based diagonal matrix designed to suppress irrelevant features, which is defined as follows:
\begin{equation}
    A_1=\operatorname{softmax}\left(W f_{\text {in }}\right),
\end{equation}
where $W$ denotes the pre-trained weight of the 1-D convolution\cite{xiang2023agca}. Thus, the adjacency matrix can be represented as:
\begin{equation}
    A = A_{0} \times A_{1}.
\end{equation}

The second part pertains to all the object feature vectors $(M, C)$, where cosine similarity is utilized to calculate the similarity between each object feature, thereby deriving the object feature adjacency matrix $A_{2}$:
\begin{equation}
\begin{aligned}
    A_{2} &= \begin{bmatrix}
0 & S_{12}  & \cdots   & S_{1j}   \\
S_{21} & 0  & \cdots   & S_{2j}  \\
\vdots & \vdots  & \ddots   & \vdots  \\
S_{i1} & S_{i2}  & \cdots\  & S_{ij}  \\
            \end{bmatrix}.\\
% S_{i j}&=\frac{S_{i j}^{\prime}}{\sum_{k=1}^M S_{i k}},\\
% S_{ij}^{\prime} &= \hat{v}_i \cdot \hat{v}_j=\sum_{d=1}^C \hat{v}_{i d} \cdot \hat{v}_{j d}.\\
\end{aligned}
\end{equation}
Where $S_{ij}^{\prime}$ represents the similarity between two object vertex in the M and $S_{ij}$ is its normalization, denoted as:
\begin{equation}
    S_{i j}=\frac{S_{i j}^{\prime}}{\sum_{k=1}^M S_{i k}^{\prime}},
    S_{ij}^{\prime} = \hat{v}_i \cdot \hat{v}_j=\sum_{d=1}^C \hat{v}_{i d} \cdot \hat{v}_{j d}.
\end{equation}
Note that when generating the adjacency matrix, self-similarity is set to zero, meaning the diagonal of $A_{2}$ is zero.

In a nutshell, the DCGA can be formulated as:
\begin{equation}
    Y = A_{2}\cdot Sigmoid(f_{in} \cdot G(f_{in},A)),
\end{equation}
where $G$ denotes the graph attention operation.

Upon inputting the $k$ pairs of object feature maps $(f_{q}^{obj}, (f_{r}^{obj})^{i})_{i \in [1,k]}$ into DCGA, M object descriptor vectors and N object descriptor vectors were obtained and denoted as $(d_{q}^{M}, (d_{r}^{N})^{i})_{i \in [1,k]}$.

\subsubsection{Object Matching Module}
\textbf{Motivation:} In the object matching problem of logical anomaly detection, the correspondences between objects in the query image and those in the reference image must adhere to certain physical constraints: i) An object in the query image may have at most one matching counterpart in the reference image; ii) An effective matching model should suppress the matching of objects that are extraneous or should not be present in the query image.

\textbf{Task Formulation}
For each pair of object descriptor vectors $(d_{i}^{M}, d_{j}^{N})_{i \in [1, M], j \in [1, N]}$, constraints i) and ii) imply that correspondences come from a partial assignment between the two sets of objects, that is, each possible correspondence should have a confidence value. Therefore, we have defined a partial soft assignment matrix $\mathbf{P} \in [0,1]^{M \times N}$ as:
\begin{equation}
\mathbf{P} \mathbf{1}_N \leq \mathbf{1}_M \quad \text { and } \quad \mathbf{P}^{\top} \mathbf{1}_M \leq \mathbf{1}_N.
\label{3_7_1}
\end{equation}
Our goal is to devise an object matching module capable of predicting the registration $\mathbf{P}$ from two sets of descriptor features.

\textbf{Optimal Matching}
An optimal transport layer \cite{sarlin2020superglue} is used to extract the object correspondences between $[1, M]$ and $[1, N]$. Specifically, we first compute a score matrix $\mathbf{S}_{i,j} \in \mathbb{R}^{M \times N}$:
\begin{equation}
\mathbf{S}_{i, j}=\left\langle d_{i}^{M}, d_{j}^{N}\right\rangle, \forall(i, j) \in M \times N,
\end{equation}
where $\left\langle \cdot,\cdot \right\rangle$ is the inner product and the feature descriptor vectors are normalized. The score matrix $\mathbf{S}_{i,j}$ is then augmented into $\overline{\mathbf{S}}_{i,j}$ by appending a new row and a new column, filled with a trash bin parameter $z$ that is pretrained in SuperGlue \cite{sarlin2020superglue}. We then utilize the Sinkhorn algorithm\cite{sinkhorn1967concerning} on $\overline{\mathbf{S}}_{i,j}$ to compute soft assignment matrix $\overline{\mathbf{P}}_{i,j}$ which is then recovered to $\mathbf{P}_{ij}$ by taking trash bin out.

Through the assignment matrix $\mathbf{P}$, we obtain the detailed matching results for the object descriptor vectors $d_{i}^{M}$, which include the index of each successfully matched descriptor vector and the corresponding index of the reference object descriptor vector to which it was matched. Through the trash bin, we acquire the indices of the query object descriptor vectors that were not matched, along with the index of the closest matching reference object descriptor vector ( In the $\mathbf{P}$ matrix, the index of the maximum value in the row corresponding to that object descriptor vector).

\subsection{Anomaly Measurement Module}
For $k$ pairs of object descriptor vectors $(d_{q}^{M}, (d_{r}^{N})^{i})_{i \in [1,k]}$, each pair is feeding into Object Matching Module, yielding $k$ assignment matrices $\mathbf{P}$ and corresponding trash bins. Based on all the $\mathbf{P}$ and trash bins, we label the object descriptor vectors $d_{i}^{M}, i \in [1,M]$ of the query image as either matched $d_{i}^{Matched}, i \in [1,Matched]$ or unmatched $d_{i}^{Unmatched}, i \in [1, Unmatched]$ ($Matched + Unmatched = M$).

OMM employs global object feature vectors for matching, making it a coarse-level semantic feature matching method suitable for logical anomaly detection. However, in real-world scenarios, structural anomalies and logical anomalies often coexist, with structural anomalies generally being subtle. In such cases, OMM demonstrates insensitivity to these minor defects, leading to erroneous matching of structurally anomalous objects with those in reference images. Consequently, relying solely on the unmatched object mask $d_{i}^{Unmatched}, i \in [1, Unmatched]$ is inadequate for computing the final anomaly map. Instead, we build an Anomaly Measurement Module (AMM) to detect anomalies for each object in the query image individually, resulting in an overall anomaly score map. 

Specifically, based on the indices from the $\mathbf{P}$ and trash bins, we calculate the difference in feature distribution for the object feature maps that were successfully matched with the corresponding maps in the $k$ reference images to create the matching score map. Similarly, we calculate the difference in feature distribution for the unmatched object feature maps with the most closely matched object feature maps in the $k$ reference images to create the non-matching score map. The final score map is obtained by adding these two score maps together. This approach allows for both the detection of logical anomalies and detailed checks for structural anomalies within each object.

\begin{figure}[htbp]
	\centering
	\includegraphics[width=\linewidth]{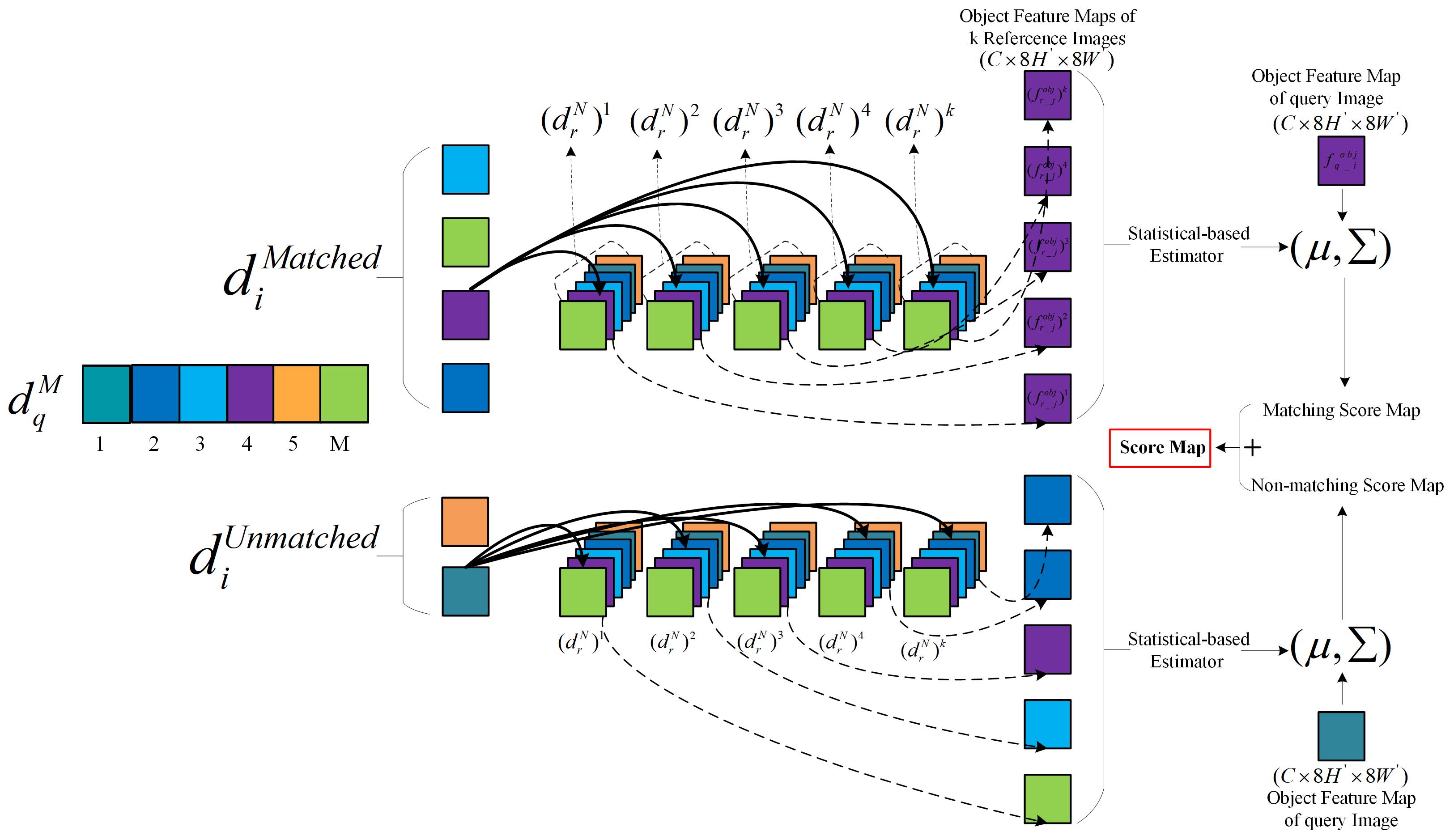}
	\caption{Architecture of the Anomaly Measurement Module (AMM).}
	\label{AMM}
\end{figure}

A statistical-based estimator is built to estimate the normal distribution of the $k$ feature maps that an object of the query image is matched with, which uses multivariate Gaussian distributions to get a probabilistic representation of the normal class. Suppose a feature map is divided into a grid of $(x,y) \in [1, H]\times[1, W]$ positions where $H \times W$ is the resolution of the feature map used to estimate the normal distribution. At each patch position $(x, y)$, let $F_{xy}=\{f_{xy}^{i},i \in [1,k]\}$ be the normal features from $k$ reference object feature maps. By the assumption that $F_{xy}$ is generated by $(\mu_{xy, \Sigma_{xy}})$, that sample covariance is:
\begin{equation}
\Sigma_{x y}=\frac{1}{k-1} \sum_{i=1}^k\left(f_{x y}^i-\mu_{x y}\right)\left(f_{x y}^i-\mu_{x y}\right)^{\mathrm{T}}+\epsilon I,
\end{equation}
where $\mu_{xy}$ is the sample mean of $F_{xy}$, and the regularization term $\epsilon I$ makes the sample covariance matrix full rank and invertible. Finally, each possible patch position is associated with a multivariate Gaussian distribution. During inference, a query object feature map that is out of the normal distribution is considered an anomaly. For a query object feature map, we use the Mahalanobis distance $\mathcal{M}(f_{xy})$ to give an anomaly score to the patch in position $(x,y)$, where
\begin{equation}
\mathcal{M}\left(f_{x y}\right)=\sqrt{\left(f_{x y}-\mu_{x y}\right)^T \Sigma_{x y}^{-1}\left(f_{x y}-\mu_{x y}\right)}.
\end{equation}
The matrix of Mahalanobis distances $\mathcal{M}=\left(\mathcal{M}\left(f_{x y}\right)\right)_{1 \leqslant x \leqslant H, 1 \leqslant y \leqslant W}$ forms an anomaly map.

For object of the $d_{i}^{Matched}, i \in [1, Matched]$, matching score map is:
\begin{equation}
    \mathcal{M}_{Matching} = \sum_{i=1}^{Matched}\mathcal{M}_{i}.
\end{equation}
For object of the $d_{i}^{Unmatched}, i \in [1, Unmatched]$, non-matching score map is:
\begin{equation}
    \mathcal{M}_{Non-matching} = \sum_{i=1}^{Unmatched}\mathcal{M}_{i}.
c\end{equation}
The final anomaly score $\mathcal{M}_{final}$ of the entire query image is:
\begin{equation}
    \mathcal{M}_{final} = \mathcal{M}_{Matching} + \mathcal{M}_{Non-matching}
\end{equation}

In a nutshell, the process of the AMM is summarized in Fig.\ref{AMM}, which is an example of the matched object and the unmatched object. For the remaining object, apply the same forward process as depicted in the example of Fig.\ref{AMM}.
\begin{table*}[htbp]
	\centering
	\caption{Comprehensive Comparison Results of the Proposed SAM-LAD and Existing Methods. Training indicates whether a method requires retraining with the dataset of a new scene when detecting anomalies in that scene. $(\uparrow)$ indicates that higher values represent better performance and $(\downarrow)$ indicates that lower values represent better performance. Bold font indicates the best results, while underlined represents the second-best results.}
	\label{tab4_1}
        \setlength{\tabcolsep}{2.1mm}
        \renewcommand{\arraystretch}{1.4}
        {\small
	\begin{tabularx}{\textwidth}{c|cccccccccc}
		\toprule
		\multirow{2}{*}{\diagbox{Methods}{Indicators}} & \multicolumn{2}{c}{Configure}     & \multicolumn{2}{c}{Capability}    & \multicolumn{2}{c}{Efficency}     & \multicolumn{2}{c}{MVTec LOCO AD}     & \multicolumn{2}{c}{MVTec AD}    \\ \cline{2-11}
		                  & Training          & Network         & Structural      & Logical         & FLOPs(Gb)$\downarrow$       & FPS$\uparrow$             & Det.$\uparrow$              & Seg.$\uparrow$              & Det.$\uparrow$           & Seg.$\uparrow$           \\ \hline
		AE\cite{bergmann2018improving}                &  $\checkmark$               & CNN             &    $\checkmark$             &     $\times$            & \textbf{5.0}             & \textbf{251.1}           & 57.4              & 37.8              & 71.0           & 80.4           \\
		f-AnoGAN\cite{schlegl2019f}          &    $\checkmark$            & CNN             &     $\checkmark$          &      $\times$           & \underline{7.7}             & \underline{133.4}           & 64.3              & 33.4              & 65.8           & 76.2           \\ \hline
		SPADE\cite{cohen2020sub}             &     $\checkmark$            & CNN             &     $\checkmark$            &    $\times$             & -               & 0.9             & 74.0              & 45.1              & 85.5           & 96.5           \\
		Padim\cite{defard2021padim}             &   $\checkmark$              & CNN             &    $\checkmark$             &     $\times$            & -               & 4.6             & 78.0              & 52.1              & 95.5           & 96.7           \\
		Patchcore\cite{roth2022towards}         &    $\checkmark$             & CNN             &    $\checkmark$             &      $\times$           & 11.4            & 25.1            & 83.5              & 34.3              & \underline{99.1}           & \underline{98.1}           \\ \hline
		THFR\cite{guo2023template}              &   $\checkmark$              & CNN             &     $\checkmark$            &     $\checkmark$            & -               & 7.69            &   \underline{86.0}                &     \underline{74.1}              &   \textbf{99.2}           &     98.2           \\
		DSKD\cite{zhang2024contextual}              &   $\checkmark$              & CNN             &      $\checkmark$           &       $\checkmark$          & -               & -               &   84.0                &    73.0               &     -           &     -           \\
		GLCF\cite{yao2023learning}              &  $\checkmark$               & Transformer     &       $\checkmark$          &    $\checkmark$             & 52.6            & 11.2            & 83.1              & 70.3              & 98.6           & 98.2           \\ \hline
		\textbf{SAM-LAD(ours)}           &   $\times$             & Transformer     &   $\checkmark$              &         $\checkmark$        &      54.7           &   8.9              &      \textbf{90.7}             &     \textbf{83.2}              &     98.4           &    \textbf{98.5}            \\ \bottomrule
	\end{tabularx}
        }
\end{table*}

\begin{figure}[t]
	\centering
	\includegraphics[width=\linewidth]{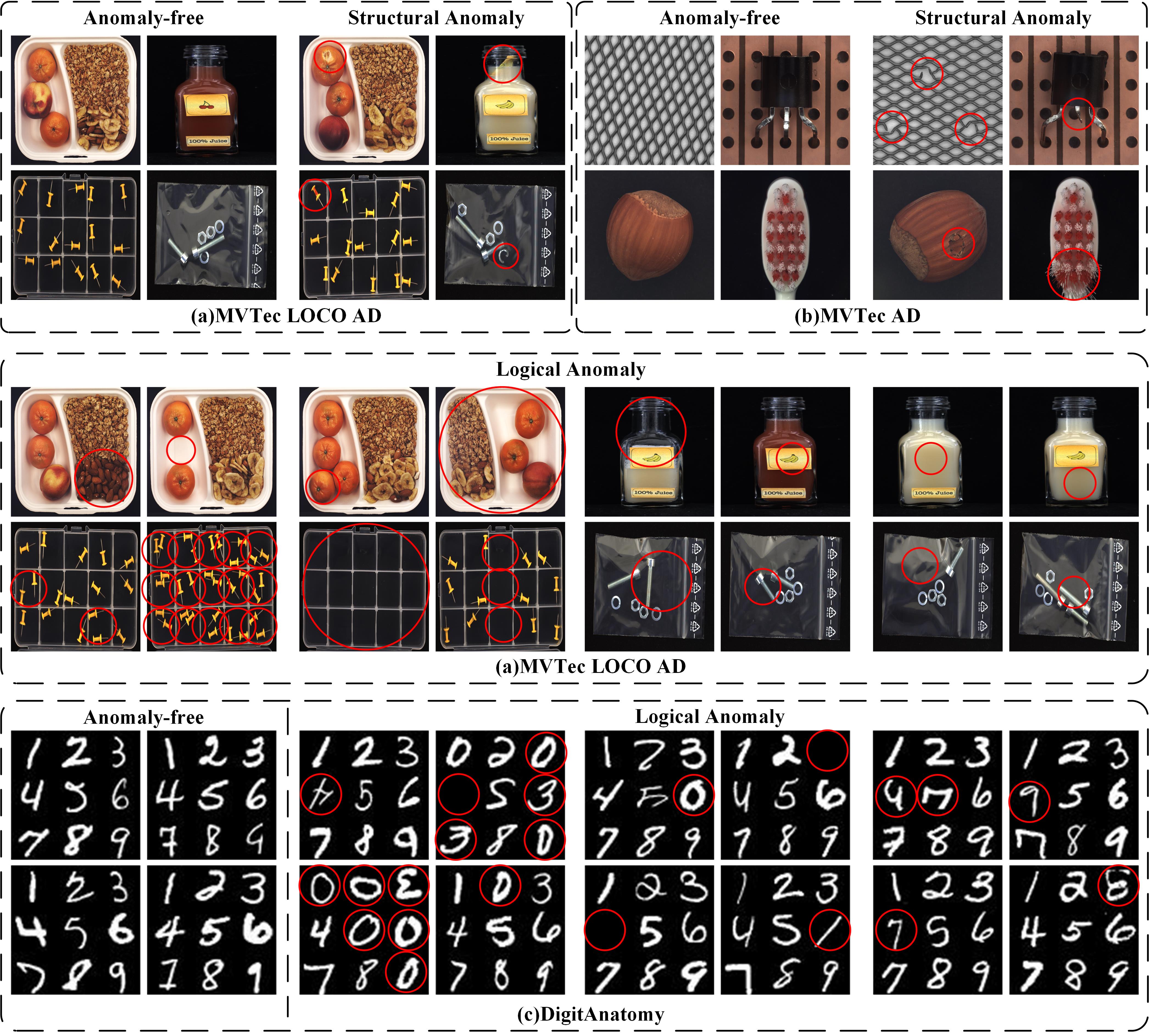}
	\caption{Datasets used in the experiments. (a) MVTec LOCO AD \cite{bergmann2022beyond}. (b) MVTec AD \cite{bergmann2019mvtec}. (c) DigitAnatomy dataset \cite{xiang2023squid}. Among these datasets, the majority of the MVTec AD consists of structural anomalies, while the MVTec LOCO AD dataset includes both logical and structural anomalies, and the DigitAnatomy dataset includes logical anomalies. The anomalies are annotated by red circles.}
	\label{4_1}
\end{figure}

\section{Experimental results}\label{sec:experiments}
In this section, we will conduct comprehensive experiments to validate the effectiveness of the proposed SAM-LAD. Specifically, we will compare it with existing methods on benchmarks from multiple scenarios and perform further analysis to verify the reasons behind the performance of the framework. Finally, we will conduct ablation studies to further analyze the framework’s performance.
\begin{table*}[t]
	\centering
        \begin{threeparttable}
	\caption{Quantitative detection and localization results of the SAM-LAD framework on the MVTec LOCO AD dataset. Results for each category are given as logical anomalies/structural anomalies or the average of both. Overall averages are given as logical anomalies/structural anomalies and the average of both. The results of the comparison methods are from \cite{bergmann2022beyond}, \cite{yao2023learning}, and \cite{xie2024iad}.}
	\label{tab4_2}
        \setlength{\tabcolsep}{1.35mm}
        \renewcommand{\arraystretch}{1.25}
        {\footnotesize
	\begin{tabularx}{\textwidth}{cccc|cccc|cccccc}
		\hline
		\multicolumn{4}{c|}{\multirow{3}{*}{Category\textbackslash Method}}                                                 & \multicolumn{4}{c|}{Baselines}& \multicolumn{5}{c}{SoTAs}&\multirowcell{3}{SAM-LAD\\(ours)}\\ \cline{5-13}
		\multicolumn{4}{c|}{}& \Methods{AE}{\cite{bergmann2018improving}}& \Methods{VAE}{\cite{kingma2013auto}}& \Methods{f-AnoGAN}{\cite{schlegl2019f}} & \Methods{MNAD}{\cite{park2020learning}}&\Methods{EfficientAD-S}{\cite{batzner2024efficientad}}&\Methods{GCAD}{\cite{bergmann2022beyond}} &\Methods{THFR}{\cite{guo2023template}} &\Methods{DSKD}{\cite{zhang2024contextual}}& \Methods{GLCF}{\cite{yao2023learning}}& \\ \hline
		\multicolumn{2}{c}{\multirowcell{7}{Image-Level\\AUROC}}& \multicolumn{2}{|c|}{Breakfast box} & 58.0/47.7& 47.3/38.3&69.4/50.7&59.9/60.2&-&87.0/80.9&78.0&-&86.7/79.1&96.7/85.2                   \\
		\multicolumn{2}{c}{}& \multicolumn{2}{|c|}{Juice bottle}&67.9/62.6&61.3/57.3&82.4/77.8&70.5/84.1&-&100/98.9&97.1&-&98.7/93.3&98.7/96.5                   \\
		\multicolumn{2}{c}{}& \multicolumn{2}{|c|}{Pushpins}&62.0/66.4&54.3/75.1&59.1/74.9&51.7/76.7&-&97.5/74.9&73.7&-&80.1/78.6&97.2/79.2                   \\
		\multicolumn{2}{c}{}& \multicolumn{2}{|c|}{Screw bag} &46.8/41.5&47.0/49.0&60.8/56.8&46.8/59.8&-&56.0/70.5&88.3&-&80.1/78.6&95.2/77.9                   \\
		\multicolumn{2}{c}{}& \multicolumn{2}{|c|}{Splicing Connectors}&56.2/64.8&59.4/54.6&68.8/63.8&57.6/73.2&-&89.7/78.3&92.7&-&89.6/89.7&91.4/88.6                   \\ \cline{3-14}
		\multicolumn{2}{c}{}                                        & \multicolumn{2}{|c|}{\multirow{2}{*}{Average}}    & \multirowcell{2}{58.2/56.6\\57.4} & \multirowcell{2}{53.8/54.8\\54.3} & \multirowcell{2}{65.9/62.7\\64.3} & \multirowcell{2}{60.1/70.2\\65.1} & \multirowcell{2}{\underline{94.1}/\underline{85.8}\\ 90.0} & \multirowcell{2}{86.0/80.7\\83.4} & \multirow{2}{*}{86.0} & \multirowcell{2}{81.2/\textbf{86.9}\\84.0} & \multirowcell{2}{82.4/83.8\\83.1} & \multirowcell{2}{\textbf{95.8}/85.5\\ \textbf{90.7}} \\
		\multicolumn{2}{c}{}& \multicolumn{2}{|c|}{}&&&&&&&&&&                   \\ \hline
		\multicolumn{2}{c}{\multirowcell{7}{Pixel-Level\\sPRO}}       & \multicolumn{2}{|c|}{Breakfast box}&18.9&16.5&22.3&8.0&-&50.2&58.3&56.8&52.8&81.9/79.1                   \\
		\multicolumn{2}{c}{}& \multicolumn{2}{|c|}{Juice bottle}&60.5&63.6&56.9&47.2&-&91.0&89.6&86.5&91.3&94.4/93.5                   \\
		\multicolumn{2}{c}{}& \multicolumn{2}{|c|}{Pushpins}&32.7&31.1&33.6&35.7&-&73.9&76.3&82.5&61.5&76.2/74.2                   \\
		\multicolumn{2}{c}{}& \multicolumn{2}{|c|}{Screw bag}&28.9&30.2&34.8&34.4&-&55.8&61.5&62.7&61.5&86.3/71.6                   \\
		\multicolumn{2}{c}{}& \multicolumn{2}{|c|}{Splicing Connectors}&47.9&49.6&19.5&44.2&-&79.8&84.8&76.7&78.5&89.1/85.2                   \\ \cline{3-14}
		\multicolumn{2}{c}{}                                        & \multicolumn{2}{|c|}{\multirow{2}{*}{Average}}    & \multirowcell{2}{46.0/29.6\\37.8} & \multirowcell{2}{45.9/30.5\\38.2} & \multirowcell{2}{46.0/20.9\\33.4} & \multirowcell{2}{26.6/41.2\\33.9} & \multirowcell{2}{\underline{74.8}/ \textbf{80.8}\\ \underline{77.8}} & \multirowcell{2}{71.1/69.2\\70.1} & \multirow{2}{*}{74.1} & \multirow{2}{*}{73.0} & \multirowcell{2}{70.0/70.6\\70.3} & \multirowcell{2}{\textbf{85.6}/ \underline{80.7}\\ \textbf{83.2}} \\
		\multicolumn{2}{c}{}& \multicolumn{2}{|c|}{}&&&&&&&&&&                   \\ \hline
	\end{tabularx}
        }
        \begin{tablenotes}
        \raggedright
        \item[1] The best performance is indicated by bold font, while the second best is indicated by an underline.
        \end{tablenotes}
        \end{threeparttable}
\end{table*}
\begin{figure*}[!t]
	\centering
	\includegraphics[width=\textwidth]{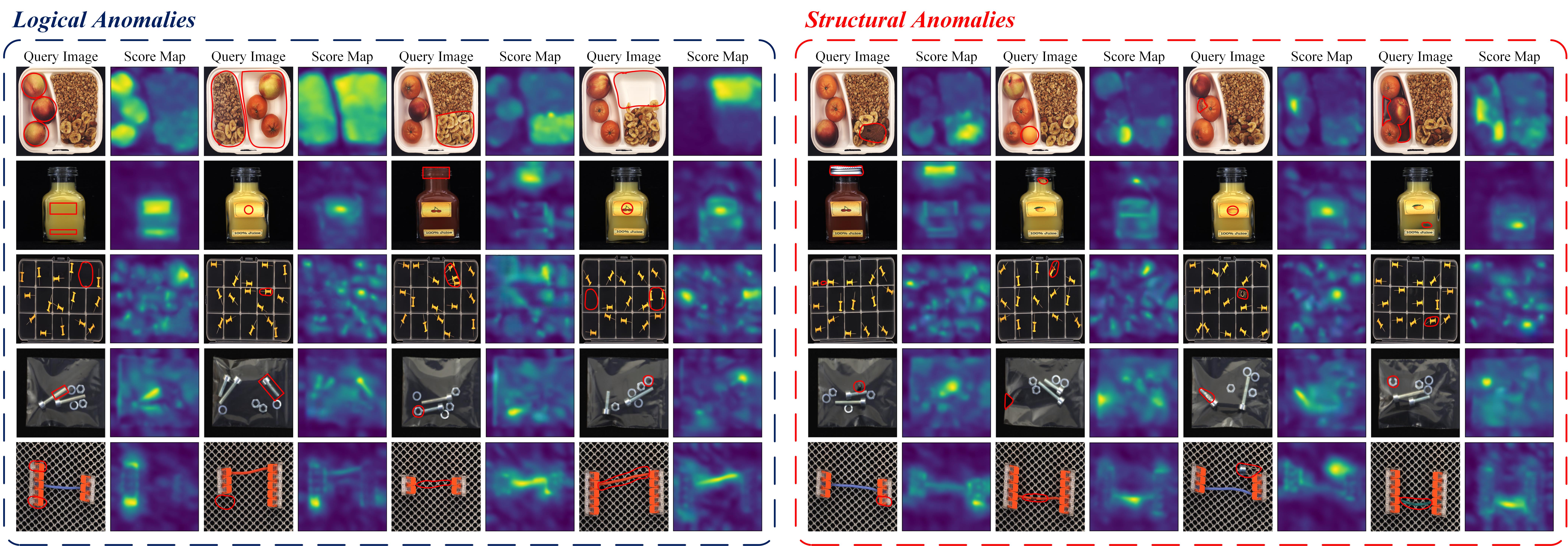}
	\caption{Examples of qualitative detection results for logical and structural anomalies using our SAM-LAD framework on the MVTec LOCO AD dataset. The Ground truth defects are annotated by red circles.}
	\label{4_3}
\end{figure*}
\subsection{Datesets}
In our experiments, we primarily use three public unsupervised anomaly detection datasets and a selection of representative samples from these datasets is illustrated in Fig.\ref{4_1}.

\textbf{MVTec LOCO AD:} The MVTec LOCO AD dataset \cite{bergmann2022beyond} was recently released by MVTec Software GmbH, which is developed explicitly for logical anomalies and comprises five object categories, each containing both structural and logical anomalies in the test set. It has a total of 2,076 anomaly-free samples and 1,568 samples for testing. Each of the 1,568 test images is either anomaly-free or contains at least one structural or logical anomaly. Specifically, In the test images containing logical anomalies, the number and types of objects are variable. In the normal setting, all objects should adhere to specific logical constraints. Logical anomalies deviate from these constraints, manifesting as missing, extra, wrong location, or inappropriate object combinations. Pixel-level annotations are provided as the ground truth for testing.

\textbf{MVTec AD:} The MVTec AD \cite{bergmann2019mvtec} comprises 10 object categories and 5 texture categories, with a total of 4,096 anomaly-free samples and 1,258 anomaly samples in the testing set. The anomaly types include only local structural damage. Pixel-level annotations are provided as the ground truth for testing.

\textbf{DigitAnatomy:} In a recent study\cite{xiang2023squid}, a groundbreaking synthetic logical dataset was introduced, consisting of digits organized in a grid pattern. Images containing digits in the correct sequential order were deemed normal, while those with deviations were categorized as abnormal. The dataset comprises a variety of simulated anomalies, including missing digits, out-of-sequence digits, flipped digits, and zero digits. These types of anomalies exhibit a greater degree of logical patterns. Image-level annotations are provided as the ground truth for testing.

\subsection{Evaluation metrics}
We use the Area Under the Receiver Operating Characteristic Curve (AUROC) score as a threshold-free metric to evaluate image-level anomaly detection. For anomaly localization in both MVTec AD and MVTec LOCO AD datasets, AUROC is also suitable for assessing structural anomalies. However, logical anomalies of the MVTec LOCO AD, e.g., a missing object, are challenging to annotate and segment on a per-pixel basis. To evaluate anomaly localization performance, we use the saturated Per-Region Overlap (sPRO) metric\cite{bergmann2022beyond} with the per-pixel false-positive rate of 5\%, which is a generalized version of the PRO metric \cite{bergmann2019mvtec}. This metric reaches saturation once it overlaps with the ground truth and achieves a predefined saturation threshold. All thresholds are also provided by the MVTec LOCO AD dataset.

\subsection{Implementation Details}
In our experiments, we resize each image to a resolution of 224 $\times$ 224 and normalize the pixel intensities based on the mean value and standard deviation obtained from the ImageNet dataset\cite{deng2009imagenet}. Additionally, the $k$-nearest is set to 2. All experiments were conducted on a computer equipped with Xeon(R) Gold 6230R CPUs@2.60GHZ and one NVIDIA A100 GPU with 40GB of memory.
\subsection{Comparison With the State-of-the-Art Models}\label{section4.4}
\begin{table*}[t]
	\centering
	\caption{The AUROC Results of Various Methods in MVTec AD at the Image/Pixel-level}
	\label{tab4_3}
        \setlength{\tabcolsep}{1.6mm}
        \renewcommand{\arraystretch}{1.1}
        {\footnotesize
	\begin{tabularx}{\textwidth}{cccccccccccc}
		\toprule
		Category\textbackslash Method      & \Methods{AE}{\cite{bergmann2018improving}}              &\Methods{PMB-AE}{\cite{xing2023visual}}               & \Methods{MKD}{\cite{salehi2021multiresolution}}                  & \Methods{RIAD}{\cite{zavrtanik2021reconstruction}}                 & \Methods{DRAEM}{\cite{bergmann2020uninformed}}                & \Methods{RD4AD}{\cite{deng2022anomaly}}                & \Methods{Padim}{\cite{defard2021padim}}                & \Methods{Patchcore}{\cite{roth2022towards}}            & \Methods{C-FLOW}{\cite{gudovskiy2022cflow}}               & \Methods{GLCF}{\cite{yao2023learning}}                 & \makecell[c]{SAM-LAD\\(ours)}        \\ \midrule
		Carpet                & 67.0/87.0            & 93.1/92.3            & 79.3/95.6            & 84.2/94.2            & 97.0/95.5            & 98.9/98.9            & 99.8/98.9            & 98.7/99.0            & 99.8/98.9            & 99.8/98.2            & 100/99.3                     \\
		Grid                 & 69.0/94.0            & 97.1/94.3            & 78.1/91.8            & 93.0/85.8            & 99.1/96.8            & 99.2/95.3            & 99.2/93.6            & 99.2/95.0            & 99.1/96.7            & 99.3/94.8            &  99.2/98.6                    \\
		Leather              & 46.0/78.0            & 94.5/96.7            & 95.1/98.1            & 100/99.4             & 100/98.6             & 100/99.4             & 100/99.1             & 100/99.3             & 100/99.1             & 100/99.0             & 100/98.9                     \\
		Tile                 & 52.0/59.0            & 97.2/90.7            & 91.6/82.8            & 98.7/89.1            & 99.6/99.2            & 99.3/95.6            & 98.1/91.2            & 98.7/95.6            & 98.1/91.2            & 99.8/95.1            &  100/96.3                    \\
		Wood                 & 83.0/73.0            & 100/86.5             & 94.3/84.8            & 99.6/96.3            & 99.9/99.3            & 100/99.3             & 96.7/94.9            & 98.2/98.7            & 96.7/94.9            & 99.7/98.9            &  97.3/94.3                    \\
		Bottle               & 88.0/93.0            & 93.7/95.2            & 99.4/96.3            & 99.9/98.4            & 99.2/99.1            & 100/98.7             & 99.9/98.1            & 100/98.6             & 100/99.0             & 100/98.4             &   100/98.6                   \\
		Cable                & 61.0/82.0            & 85.6/94.2            & 89.2/82.4            & 81.9/84.2            & 91.8/94.7            & 95.0/97.4            & 92.7/95.8            & 99.5/98.4            & 97.6/97.6            & 100/98.2             &  95.2/97.5                    \\
		Capsule              & 61.0/94.0            & 82.3./92.1           & 80.5/95.9            & 88.4/92.8            & 98.5/94.3            & 96.3/98.7            & 91.3/98.3            & 98.1/98.8            & 97.7/99.0            & 95.5/98.9            &  96.8/98.2                    \\
		Hazelnut             & 54.0/97.0            & 99.4/92.5            & 98.4/94.6            & 83.3/96.1            & 100/92.9             & 99.9/98.9            & 92.0/97.7            & 100/98.7             & 100/98.9             & 100/98.9             &  99.3/99.4                    \\
		Meta nut             & 54.0/89.0            & 85.8/84.5            & 82.7/86.4            & 88.5/92.5            & 98.7/96.3            & 100/97.3             & 98.7/96.7            & 100/98.4             & 99.3/98.6            & 100/97.8             &  99.9/98.0                    \\
		Pill                 & 60.0/91.0            & 86.1/91.1            & 82.7/89.6            & 84.5/98.8            & 93.9/97.6            & 97.0/99.6            & 85.8/97.4            & 98.1/99.4            & 91.9/98.9            & 95.3/99.4            &  98.6/98.2                    \\
		Screw                & 51.0/96.0            & 97.0/97.7            & 83.3/96.0            & 100/98.9             & 100/98.1             & 99.5/99.1            & 96.1/98.7            & 100/98.7             & 99.7/98.9            & 92.5/98.8            & 95.7/96.8                    \\
		Toothbruth           & 74.0/92.0            & 95.8/97.5            & 92.2/96.1            & 83.8/95.7            & 98.9/97.6            & 96.6/98.2            & 93.3/94.7            & 96.6/97.4            & 96.8/98.9            & 96.3/98.1            &    96.3/99.0                  \\
		Transistor           & 52.0/90.0            & 80.8/92.4            & 85.6/76.5            & 90.9/87.7            & 93.1/90.9            & 96.7/92.5            & 97.4/97.2            & 100/96.3             & 95.2/98.0            & 100/97.5             &  95.8/98.5                    \\
		Zipper               & 80.0/88.0            & 77.3/95.4            & 93.2/93.9            & 98.1/97.8            & 100/98.8             & 98.5/98.2            & 90.3/98.2            & 99.4/98.5            & 98.5/99.1            & 97.2/97.7            &  96.2/94.1                    \\ \midrule
		Average              & 63.0/87.0            & 91.8/92.1            & 87.7/90.7            & 91.7/94.2            & 98.0/97.3            & 98.4/97.8            & 95.5/96.7            & \textbf{99.1}/98.1            & 98.3/\textbf{98.6}            & 98.3/98.0            & \underline{98.4}/\underline{98.5}                     \\ \bottomrule
		\multicolumn{1}{l}{} & \multicolumn{1}{l}{} & \multicolumn{1}{l}{} & \multicolumn{1}{l}{} & \multicolumn{1}{l}{} & \multicolumn{1}{l}{} & \multicolumn{1}{l}{} & \multicolumn{1}{l}{} & \multicolumn{1}{l}{} & \multicolumn{1}{l}{} & \multicolumn{1}{l}{} & \multicolumn{1}{l}{}
	\end{tabularx}
         }
\end{table*}

In this subsection, the proposed SAM-LAD framework is analyzed in comparison with several state-of-the-art (SoTA) methods on several benchmark datasets.
\begin{figure*}[!t]
	\centering
	\includegraphics[width=\textwidth]{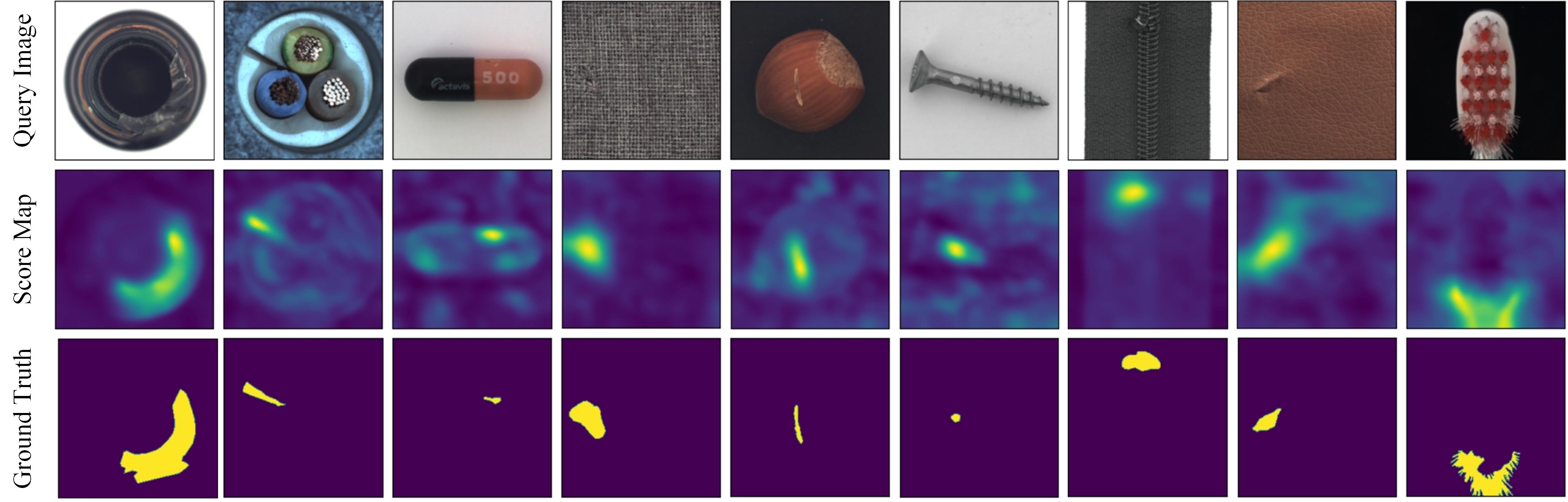}
	\caption{Examples of qualitative detection results for structural anomalies using our SAM-LAD framework on the MVTec AD dataset.}
	\label{4_6}
\end{figure*}
\subsubsection{Comprehensive Comparison} In the initial phase of our study, a comprehensive evaluation was conducted to compare the proposed SAM-LAD with existing methods, including AE\cite{bergmann2018improving}, f-AnoGan\cite{schlegl2019f}, SPADE\cite{cohen2020sub}, Padim\cite{defard2021padim}, Patchcore\cite{roth2022towards}, THFR\cite{guo2023template}, DSKD\cite{zhang2024contextual}, and GLCF\cite{yao2023learning}. The results are presented in Table \ref{tab4_1}. Notably, all existing methods require additional training for different scenarios, resulting in limited generalization. In contrast, our method achieves zero-shot capabilities, allowing plug-and-play functionality in any scene while ensuring optimal logical detection performance. Analyzing the network structures employed in these methods, it was observed that most existing approaches are built upon convolutional neural networks (CNNs), with only a few utilizing vision transformers, which are still in the early stages of exploration. A significant limitation of CNN-based models lies in their inability to capture global semantics effectively, which is crucial for logical anomaly detection.

In terms of inference efficiency and computational requirements, our SAM-LAD employs a transformer structure and upsampled feature maps, which affects the inference time compared to other schemes. However, the FLOPs of SAM-LAD are still within acceptable limits. When considering benchmark performance, the SAM-LAD demonstrates advanced capabilities in Mvtec AD for detecting structural anomalies and Mvtec LOCO AD for identifying logical anomalies, owing to its proficiency in robust segment and object matching.
\begin{table*}[t]
	\caption{Quantitative AUROC Comparison Results on the DigitAnatomy Dataset}
	\label{tab4_4}
        {\footnotesize
        \begin{tabularx}{\textwidth}{cc|ccccccccccc}
	\toprule
	\multicolumn{2}{c|}{Category\textbackslash Method}    & \Methods{AE}{\cite{bergmann2018improving}}   & \Methods{GANomaly}{\cite{akcay2019ganomaly}} & \Methods{f-AnoGAN}{\cite{schlegl2019f}} & \Methods{SQUID}{\cite{xiang2023squid}} & \Methods{Fastflow}{\cite{yu2021fastflow}} & \Methods{Patchcore}{\cite{roth2022towards}} & \Methods{DRAEM}{\cite{bergmann2020uninformed}} & \Methods{MKD}{\cite{salehi2021multiresolution}}  & \Methods{RD4AD}{\cite{deng2022anomaly}} & \Methods{GLCF}{\cite{yao2023learning}} & SAM-LAD(ours) \\ \midrule
	\multicolumn{2}{c|}{AUROC}              & 50.2 & 62.9     & 54.3     & 55.7  & 56.2     & 59.4      & 52.6  & 54.8 & 58.5  & 78.6 & \textbf{92.3}    \\ \bottomrule
        \end{tabularx}
        }
\end{table*}
\begin{figure}[t]
	\centering
	\includegraphics[width=\linewidth]{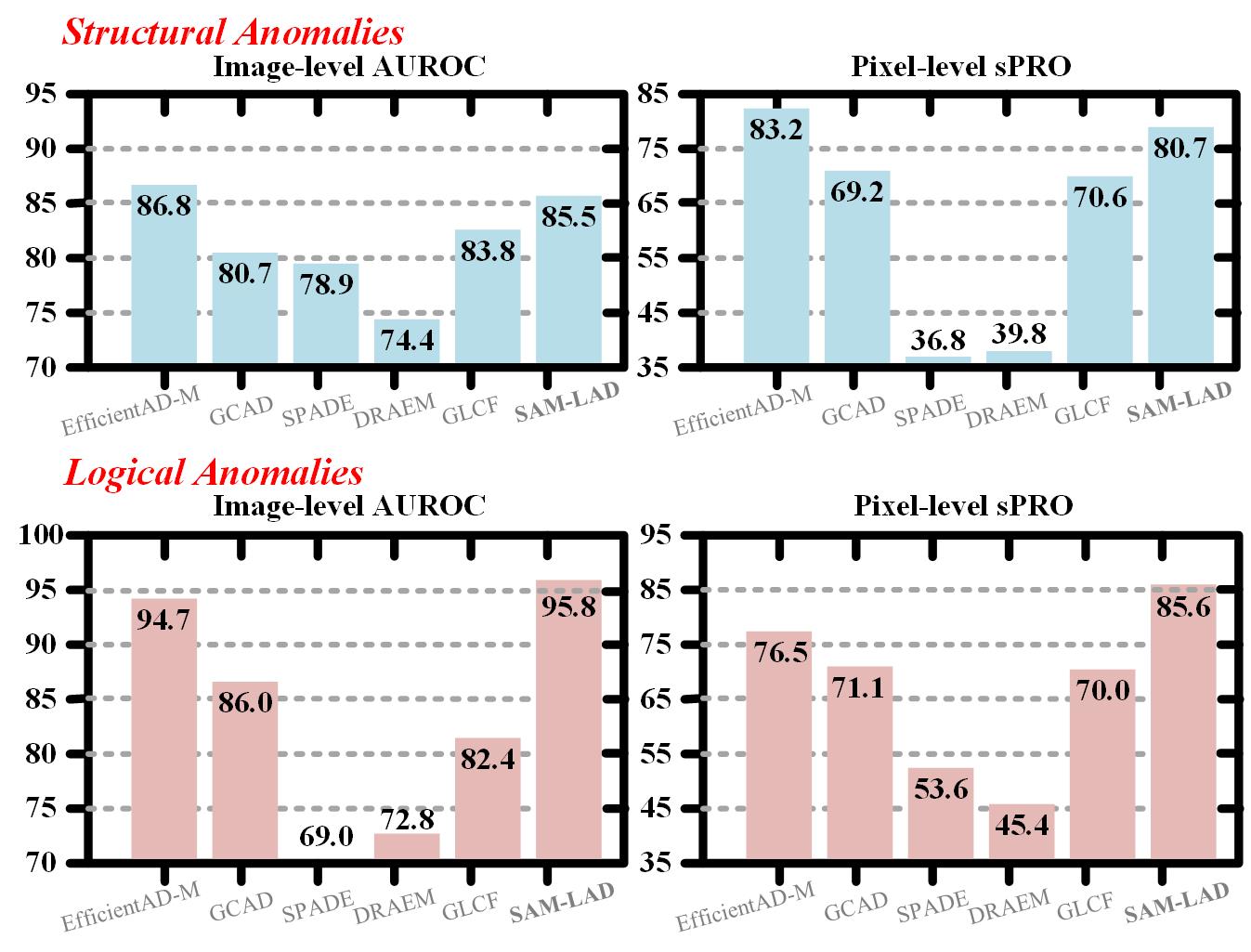}
	\caption{Comparison results of structural anomalies and logical anomalies on
MVTec LOCO AD dataset.}
	\label{4_4}
\end{figure}
\begin{figure}[t]
	\centering
	\includegraphics[width=\linewidth]{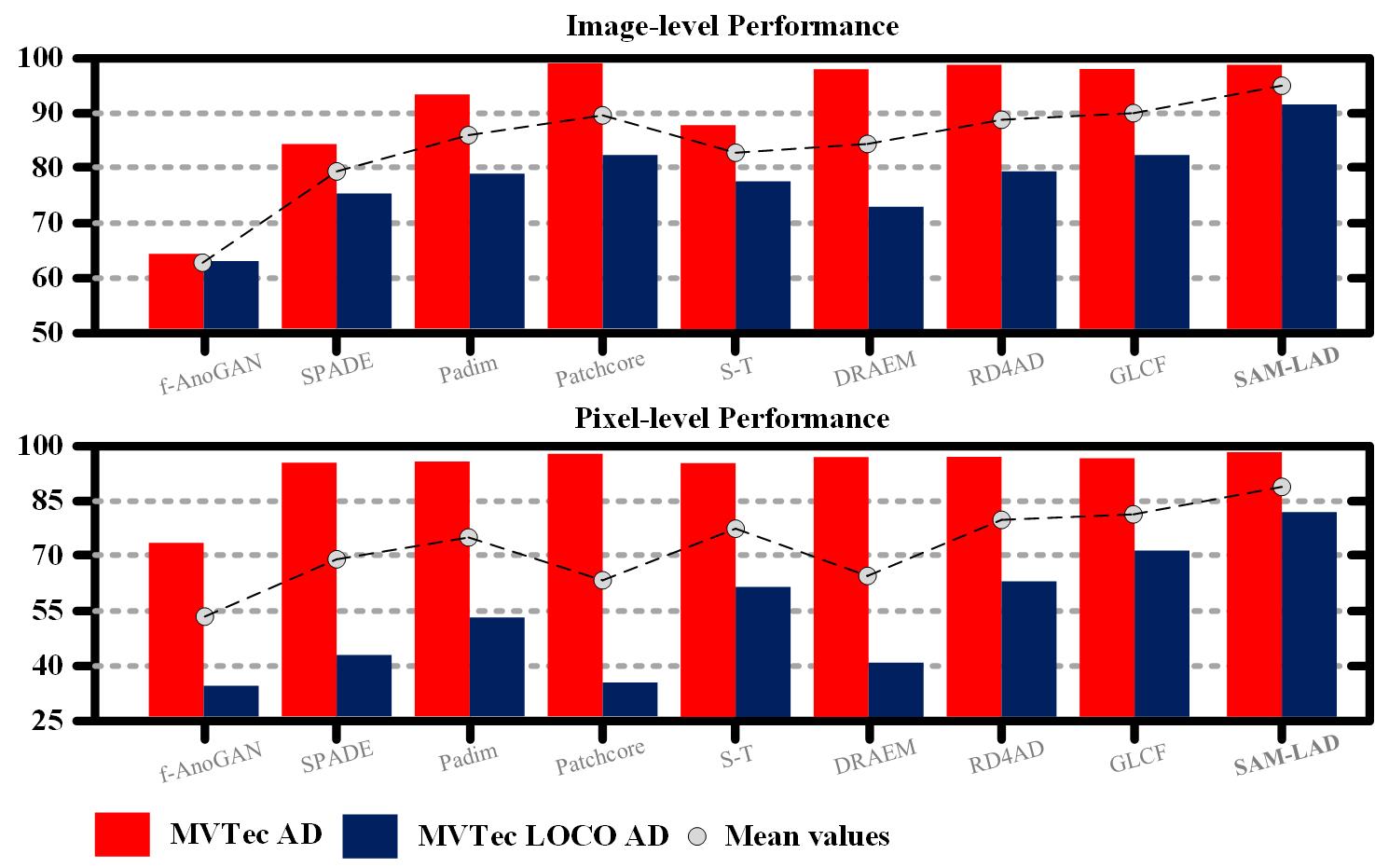}
	\caption{Image and pixel-level performance comparison of the proposed SAM-LAD and existing methods on two benchmarks}
	\label{4_5}
\end{figure}
\subsubsection{\textbf{MVTec LOCO AD}} 
We compare our proposed SAM-LAD framework with existing methods, including baseline approaches such as f-AnoGAN\cite{schlegl2019f}, AE\cite{bergmann2018improving}, VAE\cite{kingma2013auto}, and MNAD\cite{park2020learning}, as well as the top 5 best-performing SoTA methods on MVTec LOCO AD dataset's leaderboard on "Papers with Code\footnote{https://paperswithcode.com/sota/anomaly-detection-on-mvtec-loco-ad}": EfficientAD-S\cite{batzner2024efficientad}, GCAD\cite{bergmann2022beyond}, THFR\cite{guo2023template}, DSKD\cite{zhang2024contextual}, and GLCF\cite{yao2023learning}.

The comparative results are presented in Table \ref{tab4_2}. Compared to the best baseline method and the SoTA method, our framework improves by +25.6 and +0.7 on image-level AUROC, +45.0 and +5.4 on pixel-level sPRO, respectively.

Note that, among all the comparison methods, only our approach implemented zero-shot while simultaneously maintaining the best detection performance. Furthermore, Fig.\ref{4_3} displays several detection results of SAM-LAD on both structural and logical anomalies within the MVTec LOCO AD dataset, illustrating that our framework is adept at precisely pinpointing both types of anomalies.

Fig.\ref{4_4} depicts the efficacy of each method (EfficientAD-M, GCAD, SPADE\cite{cohen2020sub}, DRAEM\cite{bergmann2020uninformed}, GLCF) on the MVTec LOCO AD dataset concerning both structural and logical anomalies. The results indicate that our proposed strategy showcases strong capabilities in identifying both structural and logical anomalies. It is noteworthy that, compared to the best SoTA method, the pixel-level detection performance for logical anomalies has witnessed a significant increment of 9.1 through our framework. This highlights our method's superior proficiency in discerning logical anomalies, which aligns perfectly with our foundational intent. It enables an in-depth comprehension of the logical interrelations among each object within the entire scene and proficiently identifies the corresponding logical discrepancies.

\subsubsection{\textbf{MVTec AD}} Beyond affirming the performance of our SAM-LAD in detecting logical anomalies, we further appraise its capability to identify structural anomalies on the MVTec AD dataset, which exclusively encompasses structural anomalies. To be specific, given that each subset within the MVTec AD dataset features a scene with a single object against a regular background, we facilitate the segmentation into this single object by employing the SAM filter for subsequent matching and computation. Owing to the one-to-one fixed matching paradigm, the ultimate detection results are directly yielded by the AMM, thereby serving as an indirect assessment of the efficacy of our proposed AMM.
We compare SAM-LAD with several SoTA methods, which are AE-SSIM\cite{bergmann2018improving}, PMB-AE\cite{xing2023visual}, MKD\cite{salehi2021multiresolution}, RD4AD\cite{deng2022anomaly}, RIAD\cite{zavrtanik2021reconstruction}, DRAEM\cite{bergmann2020uninformed}, Padim\cite{defard2021padim}, Patchcore\cite{roth2022towards}, C-FLOW AD\cite{gudovskiy2022cflow}, and GLCF\cite{yao2023learning}. Table \ref{tab4_3} shows the quantitative comparison results. The proposed SAM-LAD achieves remarkable image-level anomaly detection results and pixel-level anomaly localization results, obtaining an AUROC of 98.4/98.5 across 15 categories. Remarkably, SAM-LAD exhibits significantly better performance compared to baseline methods. Moreover, compared with the SoTA methods such as Patchcore and C-FLOW, our framework achieves comparable detection and localization accuracy. A selection of qualitative results on the MVTec AD dataset is depicted in Fig.\ref{4_6}.

In conclusion, the proposed SAM-LAD showcases exceptional performance in the context of industrial anomaly detection. Fig.\ref{4_5} comprehensively compares SAM-LAD and existing methods regarding their image-level and pixel-level detection capabilities on the MVTec AD and MVTec LOCO AD datasets. The results unequivocally indicate that our framework attains the highest precision in anomaly detection and localization across these benchmarks, thereby evidencing the efficacy and versatility of our SAM-LAD in addressing an array of anomalies, including localized structural anomalies and complex logical anomalies.

\subsubsection{\textbf{DigitAnatomy}}
\begin{figure}[htbp]
    \centering
    \includegraphics[width=0.85\linewidth]{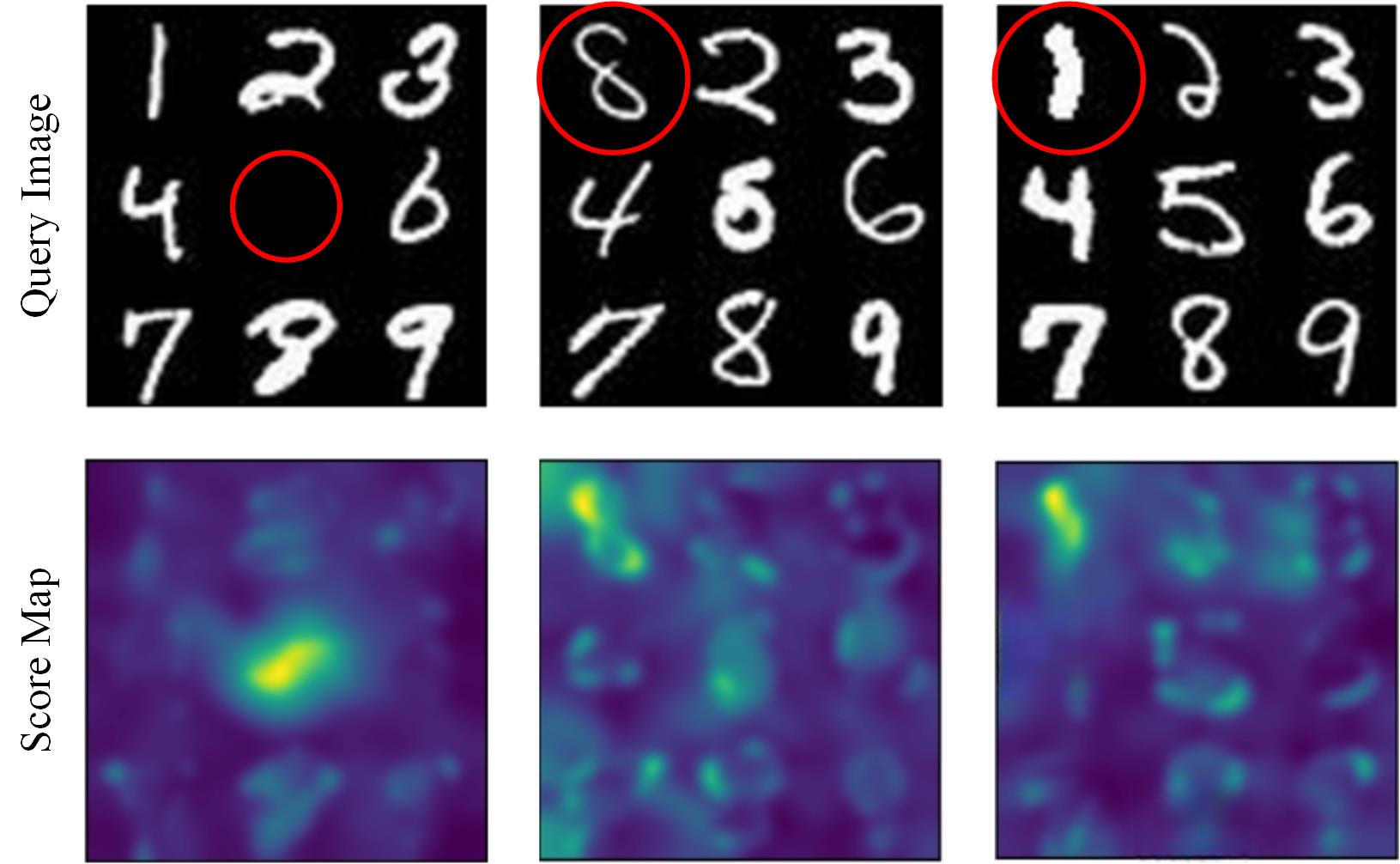}
c    \caption{Example of qualitative detection results for logical anomalies using our SAM-LAD framework on the Digitanatomy dataset. The anomalies are annotated by red circles.}
    \label{4_7}
\end{figure}

To further validate the efficacy of the proposed SAM-ALD in detecting logical anomalies, we conduct a comparative experiment employing the DigitAnatomy dataset. Since the DigitAnatomy dataset contains only logical anomalies, we can utilize a lightweight version of SAM-LAD for its detection. Specifically, the lightweight version calculates the anomaly map in AMM exclusively for non-matched objects. Thus, the Eq.15 is modified as:
\begin{equation}
    \mathcal{M}_{final} = \mathcal{M}_{Non-matching} = \sum_{i=1}^{Unmatched}\mathcal{M}_{i}.
\end{equation}
A gamut of comparative assessments was performed, utilizing an array of methods such as AE\cite{bergmann2018improving}, GANomaly\cite{akcay2019ganomaly}, f-AnoGAN\cite{schlegl2019f}, and SQUID\cite{xiang2023squid}, coupled with the integrative Fastflow\cite{yu2021fastflow} and Patchcore\cite{roth2022towards}, in conjunction with Draem\cite{bergmann2020uninformed}, MKD\cite{salehi2021multiresolution}, RD4AD\cite{deng2022anomaly}, and GLCF\cite{yao2023learning}.
The comparative results are delineated in Table \ref{tab4_4} and the qualitative results are shown in Fig.\ref{4_7}. The results show that our framework markedly transcends existing methods. In particular, the most efficacious GLCF method, our novel SAM-LAD method, realizes a substantial enhancement, accruing a gain of +13.7 in AUROC. Albeit the SQUID method was devised alongside the DigitAnatomy dataset, it garners merely an unassuming AUROC of 55.7. Conversely, our SAM-LAD framework attains a formidable AUROC of 92.3, thereby accentuating its preeminent efficacy.

\subsection{Further Analysis}
The outstanding logical anomaly detection results of our framework are based on the excellent  performance of the Object Matching Model (OMM). Therefore, to verify the reasons behind the exceptional performance, for each category in MVTec LOCO AD and DigitAnatomy dataset, we compute the precision, recall, and F1 score of OMM's matching results, which are shown in Table \ref{tab4_5}. 

The results demonstrate the OMM's strong matching capability and validate its effectiveness. However, the matching accuracy for the pushpins and splicing connectors categories drops significantly. This occurs because each object is very similar in these scenarios, posing a challenge for the OMM. Instead of using unmatched object masks as the anomaly score, we designed the AMM (Anomaly Measurement Module) to further detect anomalies. This design reduces the SAM-LAD's dependency on matching accuracy, as shown by the final anomaly detection results in Section \ref{section4.4}.
\begin{table}[htbp]
	\centering
	\caption{The Performance of the OMM's Matching capability in MVTec LOCO AD dataset and Digitanatomy Dataset.}
	\label{tab4_5}
        \setlength{\tabcolsep}{3mm}
        \renewcommand{\arraystretch}{1.3}
	\begin{tabularx}{\linewidth}{c|ccc}
		\toprule
		Category\textbackslash Metrics   & Precision & Recall & F1 Score \\ \hline
            \multicolumn{4}{c}{MVTec LOCO AD}\\ \hline
		Breakfast box       & 99.2\%     & 98.8\%  & 99.0\%    \\
		Juice Bottle        & 99.5\%     & 98.3\%  & 98.9\%    \\
		Pushpins            & 89.0\%     & 86.9\%  & 87.9\%    \\
		Screw bag           & 99.5\%     & 98.9\%  & 99.2\%    \\
		Splicing connectors & 91.1\%     & 87.9\%  & 89.5\%    \\ \hline
		Mean                & 95.7\%     & 94.2\%  & 96.5\%    \\ \hline
            \multicolumn{4}{c}{Digitanatomy Dataset}\\ \hline
            Mean                    & 96.5\%     & 95.1\%  & 95.8\%\\  \bottomrule
	\end{tabularx}
\end{table}

\subsection{Ablation Experiment}\label{section4.6}
\subsubsection{Impact of the Backbone}We use the pre-trained backbone as the feature extractor in the proposed SAM-LAD. In this study, we primarily considered three different pre-trained backbone networks: the Wide-ResNet50 of the CNN type, the Swin-transformer of the ViT type, and the DINOv2 of the ViT type. The main results are presented in Table \ref{tab4_6}, where it is clear  that DINOv2 manifests the most superior performance. Therefore, we adopt DINOv2 as the feature extractor for the proposed framework.
\begin{table}[htbp]
    \centering
    \caption{Ablation Experiments of the Different Backbone on the MVTec LOCO AD Dataset.}
    \label{tab4_6}
    {\footnotesize
    \setlength{\tabcolsep}{2.5mm}
    \begin{tabularx}{\linewidth}{c|c|c|c}
    \toprule
       Metrics\textbackslash Network  &  Wide-ResNet50 & Swin-Transformer & DINOv2 \\
       \hline
         Det.(AUC) & 82.3 & 85.8 & \textbf{90.7} \\
         Seg.(sPRO) & 72.9  & 79.7 & \textbf{83.2}\\
    \bottomrule
    \end{tabularx}
    }
\end{table}
\subsubsection{Impact of the FeatUp Configuration}
\begin{figure}[htbp]
    \centering
    \includegraphics[width=0.9\linewidth]{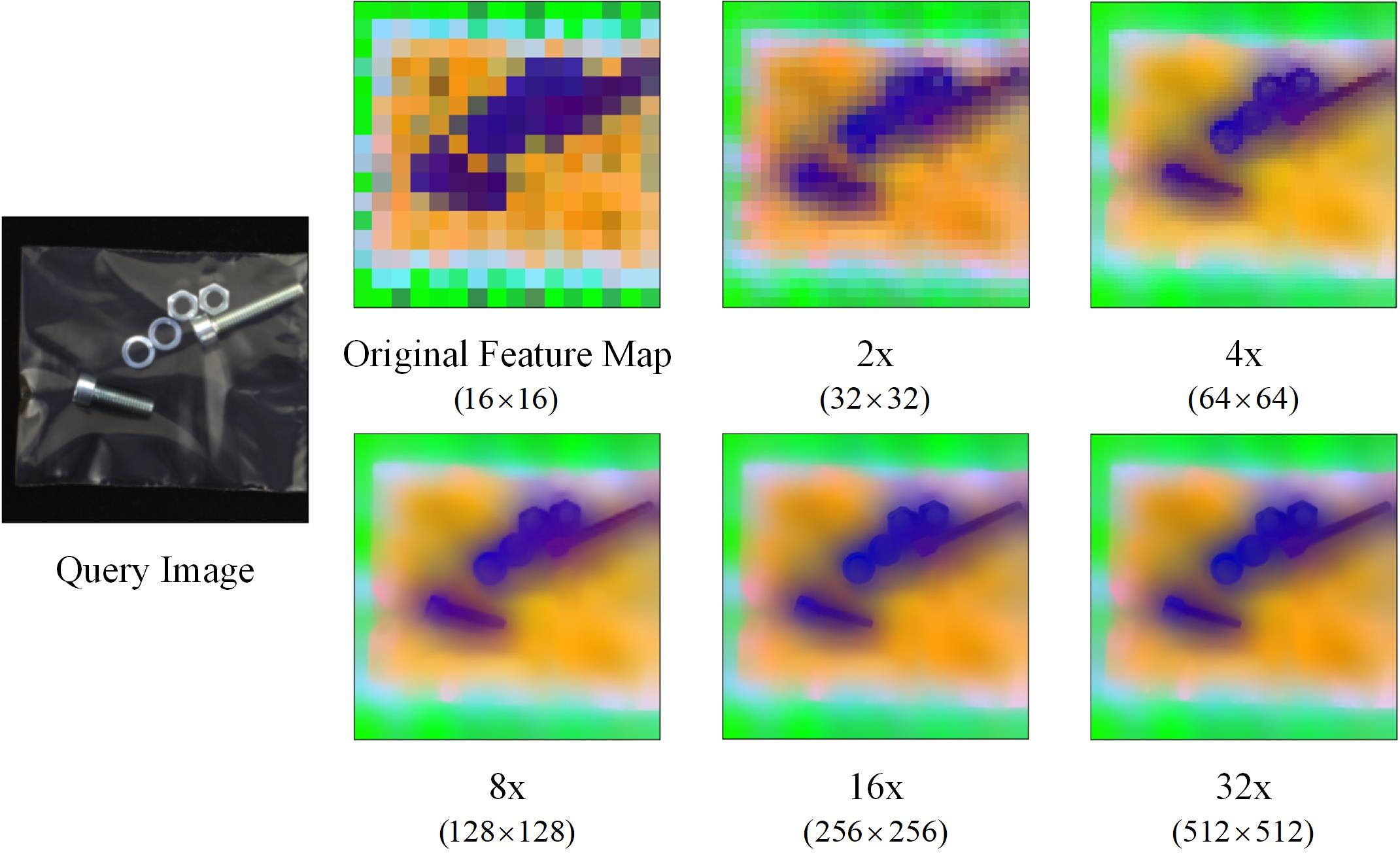}
    \caption{Visualizing PCA components with different upsampling configurations of the FeatUp in MVTec LOCO AD.}
    \label{4_8}
\end{figure}

After obtaining the feature maps from the images, we use the FeatUp operation to upsample these maps and restore the lost spatial information. In the study of FeatUp\cite{fu2024featup}, the authors proposed five upsampling configurations: 2$\times$, 4$\times$, 8$\times$, 16$\times$, and 32$\times$, shown in Fig.\ref{4_8}. To verify FeatUp's efficacy and assess the impact of different upsampling configurations on anomaly detection performance, we conduct an ablation study on the upsampling factors.
The main results are shown in Table \ref{tab4_7}. Compared to the original feature maps, FeatUp significantly enhances detection capabilities. Specifically, as the upsampling factor increases, SAM-LAD's detection and segmentation performance improves. However, upsampling the feature maps by 16$\times$ and 32$\times$ significantly  increases the model's FLOPs and drastically reduces inference speed, which is impractical for industrial use. Therefore, to balance detection performance with real-time inference needs, we chose an 8$\times$ upsampler to restore lost spatial information in the feature maps for subsequent computations.

\begin{table}[htbp]
        \centering
	\caption{Ablation experiments of the Different FeatUp's Upsampling Factors on the MVTec LOCO AD Dataset.}
	\label{tab4_7}
        \setlength{\tabcolsep}{1mm}
        \renewcommand{\arraystretch}{1.3}
        \begin{tabular}{cc|cccccc}
	\hline
	\multicolumn{2}{c|}{Metrics\textbackslash Factors}& 1$\times$ & 2$\times$ & 4$\times$ & 8$\times$ & 16$\times$ & 32$\times$ \\ \hline
	\multirow{2}{*}{\textbf{Performance}} & \multicolumn{1}{|c|}{Det.(AUC)}& 81.4 & 83.1 & 86.2 & 90.7 & 91.2 & 91.8\\
	& \multicolumn{1}{|c|}{Seg.(sPRO)}& 71.1 & 74.1 & 79.7 & 83.2& 83.8 & 84.3\\ \hline
	\multicolumn{1}{c}{\multirow{2}{*}{\textbf{Efficency}}} & \multicolumn{1}{|c|}{FLOPs(Gb)}& 9.2 & 15.2 & 23.8 & 54.7 & 165.1 & 294.2\\
	& \multicolumn{1}{|c|}{FPS} & 87.2 & 45.5 & 21.4 & 8.9 & 2.9 & 1.1 \\ \hline
        \end{tabular}
\end{table}
\begin{table}[!ht]
        \centering
	\caption{Ablation Experiments of the Different Feature Compress Methods on the MVTec LOCO AD Dataset.}
	\label{tab4_8}
        \setlength{\tabcolsep}{1mm}
        \renewcommand{\arraystretch}{1.2}
        \begin{tabular}{cc|ccc}
	\hline
	\multicolumn{2}{c|}{Metrics\textbackslash Methods}& GAP & GMP & DCGA \\ \hline
	\multirow{3}{*}{\textbf{Object Matching}} & \multicolumn{1}{|c|}{Precision}  & 90.65\% & 92.26\% & \textbf{95.66\%} \\
        & \multicolumn{1}{|c|}{Recall}   &   89.02\%   &  91.43\%   &  \textbf{94.16\%} \\
	& \multicolumn{1}{|c|}{F1 Score}   &   89.83\%  &   91.84\%  &  \textbf{96.52\%}  \\ \hline
	\multicolumn{1}{l}{\multirow{2}{*}{\textbf{Anomaly Detecting}}} & \multicolumn{1}{|c|}{Det.(AUC)}  &  85.3   &  86.2   &   \textbf{90.7}  \\
	& \multicolumn{1}{|l|}{Seg.(sPRO)} &  74.7   &  76.2   &   \textbf{83.2}   \\ \hline
        \end{tabular}
\end{table}
\subsubsection{Impact of the DCGA}
To validate the effectiveness of DCGA, we conducted an ablation experiment. DCGA was designed to extract high-dimensional features into a single vector, simplifying the similarity computation between objects. Pooling is the most widely used method for feature extraction. We compared DCGA with global average pooling (GAP) and global max pooling (GMP), as shown in Table \ref{tab4_8}. GMP outperforms GAP because only certain areas of the object feature map have discernible values, while the background is primarily null. Therefore, GMP is more effective in capturing object features in such scenarios. However, simply selecting the maximum value for pooling is insufficient. The introduction of DCGA greatly enriched the description of object features. Using graph neural network principles, DCGA extracts compelling object features from high-dimensional channels and captures feature interrelationships among objects in the same scene. This substantially improves the matching capabilities of the OMM, confirming the validity of DCGA.

\section{Discussion}
\begin{figure}[htbp]
    \centering
    \includegraphics[width=0.9\linewidth]{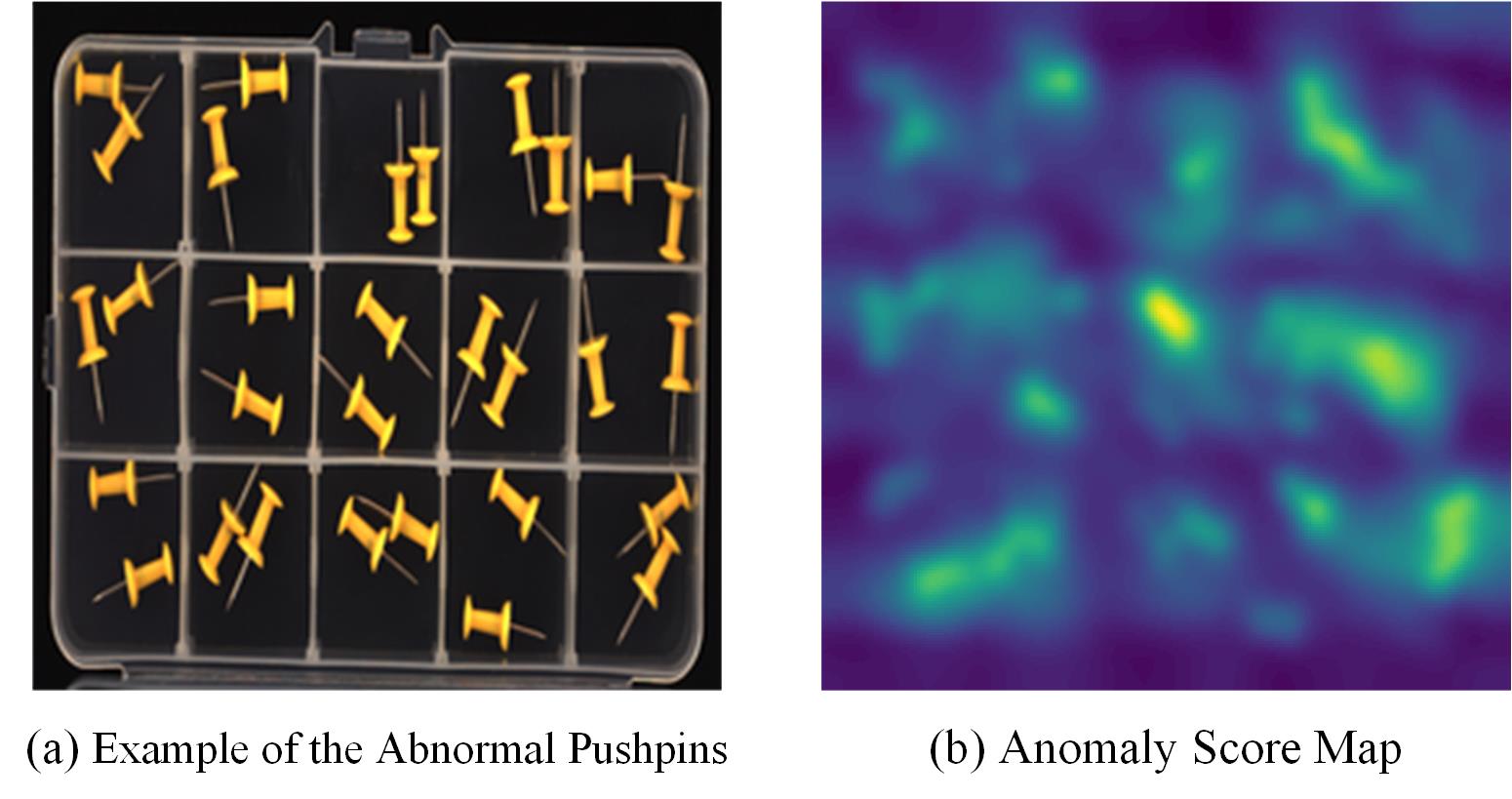}
    \caption{Failure case of SAM-LAD in MVTec LOCO AD.}
    \label{5_1}
\end{figure}

Within the scope of our investigation, we have identified a limitation that may hinder the overall performance of SAM-LAD. Specifically, the SAM-LAD encounters limitations when multiple objects of the same category are present in the scene to be inspected. This limitation stems from the fact that the core of SAM-LAD relies on an explicit matching principle to identify unmatched objects and detect anomalies. For instance, in the case of the logical anomaly in the box of pushpins shown in Fig.\ref{5_1}(a), where each compartment is expected to contain only one pushpin, SAM-LAD is expected to detect the abnormality of having two pushpins in each compartment. Unfortunately, due to the high similarity of all pushpin features in the scene, the OMM of SAM-LAD fails to accurately identify the extra pushpins in each compartment when compared with the normal reference image. As a result, as shown in Fig.\ref{5_1}(b), SAM-LAD fails to detect the logical anomaly in this scenario. The core issue lies in SAM's difficulty with segmentation granularity, often resulting in outputs that are either overly fine or too coarse.

Future research could integrate state-of-the-art visual foundation models and clustering techniques to achieve precise segmentation of key contextual objects in a scene. Specifically, the Recognize Anything Model\cite{zhang2024recognize} could be employed to identify objects in the scene and generate corresponding labels. Subsequently, the Grounded SAM\cite{ren2024grounded} method generates masks for all detected elements. If Grounded SAM produces multiple masks, indicating the presence of multiple objects, clustering methods will be applied to refine these masks. Finally, the refined segmentation mask of each key object is passed to OMM. Even for objects of the same type, OMM can accurately match them due to the refined segmentation map.

Another future promising approach is to fine-tune SAM within a specific domain to enhance its segmentation performance in targeted application scenarios, such as industrial inspection. Although SAM is a segmentation model with robust generalization capabilities, its performance can still be further optimized for specific domains through fine-tuning. Specifically, industrial components often exhibit repetitive patterns, distinct textures, and unique visual features that differ significantly from the general datasets used to train SAM. Fine-tuning domain-specific data can enable SAM to adapt to these features, thereby achieving more accurate and consistent segmentation results. This directly addresses the challenge where SAM-LAD struggles to accurately segment highly similar objects in a scene, which weakens anomaly detection performance. Additionally, industrial inspection often involves controlled yet varying lighting conditions. A fine-tuned SAM model can adapt to these conditions and mitigate the impact of lighting variations on segmentation performance.
\section{Conclusion}
In this paper, we novelty propose a zero-shot framework called SAM-LAD to address logical anomaly detection in complex scenes. We introduce a pre-trained SAM to obtain masks for all objects in the query image. Utilizing the pre-trained DINOv2 and FeatUp operations, we derive the upsampled feature map. Through the imageNNS on the query image, we obtain a reference image and its corresponding upsampled feature map. By sequentially combining object masks with upsampled feature maps, we acquire each object feature map in both the query and reference images. Subsequently, we regarded each object as a key point and employed the proposed DCGA mechanism to efficiently compress each object's feature map into a feature vector. Then, we propose the OMM, matching all object feature vectors from the query image with those in the reference image to obtain a matching matrix. Finally, based on the matching matrix, we propose AMM to compute the distribution difference estimation of the matched objects' features, resulting in the final anomaly score map.